\pdfoutput=1

\documentclass[11pt]{article}

\usepackage{emnlp2021}

\usepackage{times}
\usepackage{latexsym}
\usepackage{makecell}
\usepackage{numprint}
\usepackage[T1]{fontenc}

\usepackage[utf8]{inputenc}
\usepackage{hyperref}

\usepackage{microtype}
\usepackage{multirow}
\usepackage{graphicx}
\usepackage{array}
\newcolumntype{H}{>{\setbox0=\hbox\bgroup}c<{\egroup}@{}}
\usepackage{float}
\usepackage{enumitem}

\newcounter{mycounter}

\newcommand\labelledmodelcounter[1]{\refstepcounter{mycounter}\textbf{\themycounter}\label{model:#1}}
\newcommand{\modelref}[1]{\textbf{\ref{model:#1}}}

\title{Efficient Inference for Multilingual Neural Machine Translation}

\author{Alexandre B\'erard\thanks{~~\texttt{first.last@naverlabs.com}} \\
  NAVER LABS Europe \\\And
  Dain Lee\thanks{~~\texttt{dain.l@navercorp.com}} \\
  NAVER Corp. \\\And
  St\'ephane Clinchant$^\star$ \\
  NAVER LABS Europe \\\AND
  Kweonwoo Jung\thanks{~~\texttt{kweonwoo.jung@navercorp.com}} \\
  NAVER Corp. \\\And
  Vassilina Nikoulina$^\star$ \\
  NAVER LABS Europe \\
}

\begin{document}
\maketitle
\begin{abstract}

Multilingual NMT has become an attractive solution for MT deployment in production. But to match bilingual quality, it comes at the cost of larger and slower models. In this work, we consider several ways to make multilingual NMT faster at inference without degrading its quality. We experiment with several ``light decoder" architectures in two 20-language multi-parallel settings: small-scale on TED Talks and large-scale on ParaCrawl. Our experiments demonstrate that combining a shallow decoder with vocabulary filtering leads to more than $\times$2 faster inference with no loss in translation quality. We validate our findings with BLEU and chrF (on 380 language pairs), robustness evaluation and human evaluation.
\end{abstract}

\section{Introduction}

Multilingual machine translation \cite{johnson-etal-2017-googles, bapna-firat-2019-simple, aharoni-etal-2019-massively, zhang-etal-2020-improving,fan2020beyond,lyu-etal-2020-revisiting} has made a lot of progress in the last years. It is attractive because it allows handling multiple language directions within a single model, thus significantly reducing training and maintenance costs. However, to preserve good performance across all the language pairs, both the vocabulary size and model size have to be increased compared to bilingual NMT, which hurts inference speed. For example, the recently released M2M-100 \cite{fan2020beyond} has 15B parameters and needs multiple GPUs for inference. 
The problem of inference speed has been well studied in bilingual settings \cite{kasai2021deep, kasai2021finetuning, chen2018best,hsu-etal-2020-efficient,kim-etal-2019-research,li-etal-2020-shallow, shi-knight-2017-speeding}. One line of work consists in using \textit{lighter decoder} architectures (e.g., shallow decoder: \citealp{kasai2021deep}, RNN decoder: \citealp{chen2018best,kim-etal-2019-research, lei2021attention}). These works demonstrate that it is possible to significantly speed up inference with almost no loss in translation quality as measured by BLEU.

\begin{figure}[th!]
    \centering
    \includegraphics[scale=0.38]{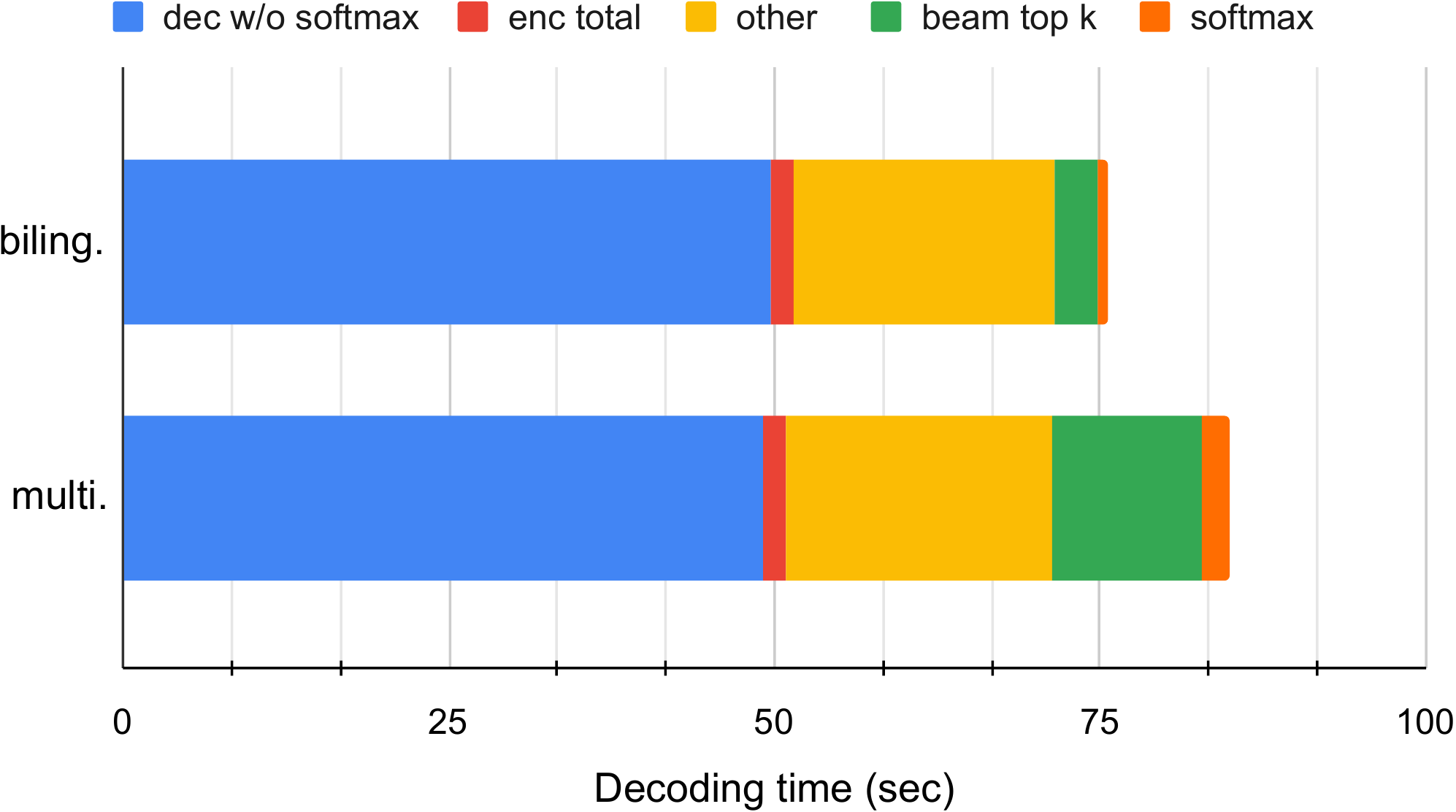}
    \caption{Decoding time spent in different components of bilingual (vocab size 16k) vs multilingual (vocab size 69k) Transformer Big 6-6 models on TED2020 valid EN-DE (average over 10 runs).}
    \label{fig:inf_speed}
\end{figure}

Figure \ref{fig:inf_speed} compares the inference time spent in each NMT component by bilingual and multilingual models of the same architecture but with different vocabulary sizes. The decoder is also the bottleneck in the multilingual model which suggests that we can expect similar speed gains with a lighter decoder. It also indicates that some speed gain could be obtained by reducing vocabulary size (which impacts both beam search and softmax).

However, it is not so obvious that lighter decoder architectures would preserve translation quality in multilingual settings, where the decoder may need more capacity to deal with multiple languages. Therefore, the goal of this work is to benchmark different architectures in terms of \textit{inference speed/translation quality} trade-off and to identify the best combination for multilingual NMT. The contributions of this paper are:
\vspace{-.15cm}
\begin{itemize}[leftmargin=1em]
    \itemsep-.25em
    \item A benchmark of two popular ``light decoder'' NMT architectures (deep encoder / shallow decoder, \citealp{kasai2021deep}; and RNN decoder, \citealp{chen2018best}) on two multilingual datasets (TED Talks and ParaCrawl) in both English-centric and multi-parallel settings. It demonstrates that the previous findings transfer to multilingual models.
    \item A combination of shallow decoder with per-language vocabulary filtering for further speed gains (achieving a global $2$~to~$3\times$ speed-up over the baseline) with no loss in translation quality.
    \item Experiments with separate language-specific shallow decoders, which trade memory for higher BLEU performance, with comparable speed as the single-decoder approach.
    \item A validation of these findings through extensive analysis, including robustness evaluation and human evaluation.
\end{itemize}

\section{Related work}
\paragraph{Lightweight decoder.} As shown in Figure~\ref{fig:inf_speed}, more than half of the inference time is devoted to the decoder and 30$\times$ more time is spent in the decoder than in the encoder (due to the autoregressive nature of the models). This explains why many efficient NMT works focus on lightweight alternatives to the Transformer decoder.
\citet{kim-etal-2019-research} perform an extensive study of various lightweight RNN architectures and obtain a $4\times$ gain in inference speed. 
\citet{kasai2021deep} show that, in bilingual settings, Transformer models with a deep encoder and shallow decoder (e.g., 10-2) can achieve similar BLEU performance as baseline 6-6 Transformers, while being much faster at inference time (on a par with current non-autoregressive MT approaches).
\citet{behnke-heafield-2020-losing} show that it is possible to prune up to 75\% of the attention heads in a Transformer, thus increasing inference speed by 50\%. Similarly, \citet{hsu-etal-2020-efficient} reduce the cost of cross-attention and self-attention by replacing it with an RNN or by pruning attention heads, obtaining up to 35\% higher speed.

Although most of the above works report speed improvements with similar BLEU scores as Transformer baselines, it is uncertain that the same will hold in multilingual many-to-many settings, where the decoder may need more capacity to deal with multiple languages. In particular, \citet{kong-etal-2021-multilingual} observe that single shallow decoders degrade one-to-many MT quality and propose to train shallow language-specific decoders or decoders that are specific to a language family or group of languages. 
    
\paragraph{Modular multilingual NMT.}
\citet{lyu-etal-2020-revisiting,escolano2020multilingual} propose modular MT models with jointly trained language-specific encoders and decoders. Such models have higher per-language capacity, increasing their performance without hurting inference speed (contrary to the common approach of training bigger multilingual models). They are also more flexible for adding new languages.
\citet{zhang2021share} study how language-specific and language-independent parameters naturally emerge in multilingual NMT. Their findings indicate that language-independent parameters can be distributed within the encoder and decoder and benefit final NMT performance. 


\section{Inference speed-up methods}
\label{sec:techniques}

\subsection{Deep encoder, shallow decoder}
\label{sec:shallow_dec}
First, we analyze how deep encoder / shallow decoder models \citet{kasai2021deep} behave in multilingual settings (many-to-many English-centric and multi-parallel). 

Our initial experiments in bilingual settings showed that a 12-2 architecture gives the best BLEU/speed trade-off (also reported by \citealp{li2021efficient}). We thus focus on this architecture and compare it with 6-6 and 6-2 architectures.

We find that in some cases (with Transformer Base on TED Talks), \textit{post-norm} 12-2 models\footnote{We prefer post-norm Transformers, as \citet{liu-etal-2020-understanding} show that when they do converge, they often lead to better performance than pre-norm.} fail to converge when trained from scratch.
When this happens, we initialize the 12-2 model with a pre-trained 6-6 model's parameters, by duplicating its encoder layers and taking its bottom 2 decoder layers. See Table~\ref{tab:deep_training_strategy} in Appendix for a comparison between this approach, training from scratch, and pre-norm Transformers.

\subsection{RNN decoder}

\citet{chen2018best} first introduced a hybrid model combining a Transformer encoder with an RNN decoder.
Hybrid Transformer/RNN models are considered a good practical choice in production settings due to their ideal performance-speed trade-off \cite{googleaiblog_rnndecoder}.
However, \citet{chen2018best} do not experiment with hybrid models in a multilingual setting, nor do they try shallower RNN decoders.
We experiment with 12-layer Transformer encoders combined with either 2-layer or 3-layer LSTM decoders (noted Hybrid 12-2 / 12-3).\footnote{Hybrid 12-3 has a similar amount of parameters as Transformer 12-2.} Because LSTMs are slower to train, we first train 12-2 Transformers which we fine-tune into Hybrid models (by initializing the LSTM decoder at random). Precise architecture details are given in Appendix~\ref{appendix:hyperparameters}.

\subsection{Target vocabulary filtering}
\label{sec:vocab_filtering}

As illustrated by Figure~\ref{fig:inf_speed}, decoding speed can also be impacted by the size of the target vocabulary, because the softmax layer's complexity is linear with respect to the vocabulary size. Some solutions have been proposed to compress vocabulary in bilingual settings: vocabulary hashing or vocabulary shortlists \cite{shu2017compressing,shi-knight-2017-speeding,senellart-etal-2018-opennmt,kim-etal-2019-research}. \citet{ding-etal-2019-call} also showed that the BPE size can be reduced drastically without hurting BLEU. However, reducing the BPE size too aggressively will result in longer sequences and hurt decoding speed.
\citet{lyu-etal-2020-revisiting} train a separate smaller BPE model per language. However, we think that this may hurt transfer learning between languages that share words (one of the reasons why multilingual NMT uses shared vocabularies in the first place).
Therefore, we propose a solution that combines the best of both worlds: have a large shared BPE vocabulary at train time, which we decompose into smaller language-specific vocabularies at test time, based on per-language token frequencies. More precisely, we train a shared BPE model of size 64k, then for each language:
\begin{enumerate}[leftmargin=1em]
    \itemsep-.1em
    \item We tokenize its training data and count the wordpiece and character occurrences.
    \item We build a vocabulary containing only tokens whose frequency is above threshold $K$ and only the $N$ most frequent wordpieces.
    \item At test time, we can filter the model's target vocabulary and embeddings to only contain these tokens, resulting in a model with a single shared source embedding matrix and several smaller per-language target embedding matrices. We call this approach ``test-time BPE filtering'' (with parameters $N_{test}$ and $K_{test}$). Appendix Tables~\ref{tab:ted_model_sizes} \&~\ref{tab:paracrawl_model_sizes} give the incurring parameter cost.
    \item We also try combining this approach with ``train-time BPE filtering'' (with parameters $N_{train}$ and $K_{train}$). For target-side training sentences in this language, we force the shared BPE model to only generate wordpieces that belong to this language's filtered vocabulary.\footnote{By using \href{https://github.com/rsennrich/subword-nmt}{\texttt{subword-nmt}}'s \texttt{-{}-vocabulary-\\threshold} option.}
\end{enumerate}

\subsection{Shared encoder, language-specific decoders}

\citet{lyu-etal-2020-revisiting} show that one can significantly increase the capacity (and thus performance) of a multilingual model without hurting decoding speed by training language-specific encoders and decoders (i.e., trading away memory for speed). 
We take the approach of a deeper shared encoder and multiple language-specific shallow decoders (similar to \citealp{kong-etal-2021-multilingual}). 
This approach keeps the memory usage to a reasonable value,\footnote{As shown in Appendix (Table~\ref{tab:paracrawl_model_sizes}) a 20-language Big 12-2 multi-decoder model has 823M parameters in total, while a Big 6-6 or Big 12-2 multi-encoder + multi-decoder model would have $\approx$20$\times$180M = 3.6B parameters.} and can maximize transfer learning on the encoder side. 

Contrary to \citet{lyu-etal-2020-revisiting} and \citet{kong-etal-2021-multilingual}, to save computation time, we first train shared multilingual MT models, which we use as initialization to our multi-decoder models (i.e., the same 2-layer decoder is copied).
We use language-specific target embeddings that are initialized with the shared embeddings obtained with the ``train-time BPE filtering'' technique described in the previous section.  
We refer to the models with shallow language-specific decoders as ``multi-decoder models.''

\subsection{Summary of notations}
\begin{itemize}[leftmargin=1em]
    \itemsep0em
    \item \textbf{Base/Big 6-6/12-2} correspond respectively to Transformer Base/Big with 6 encoder layers and 6 decoder layers (resp. 12 and 2 layers).
    \item By default, models are trained on \textbf{English-centric} data (i.e., data in all languages paired with English, in both directions).
    \item \textbf{Multi-parallel} models are fine-tuned on  data in all language directions (not just paired with English).
    \item Some models use test-time \textbf{BPE filtering} ($N_{test}$ or $K_{test}$) while others use both train-time and test-time filtering ($N_{train}$ or $K_{train}$).
    \item \textbf{Hybrid} models have a Transformer encoder and LSTM decoder and are fine-tuned from English-centric Transformers with multi-parallel data.
    \item \textbf{Multi-decoder} models have language-specific shallow decoders and are fine-tuned from English-centric models with multi-parallel data.
\end{itemize}

\section{TED Talks Experiments}
\label{sec:ted}

\subsection{Data and hyper-parameters}

We experiment with the TED Talks corpus \cite{qi-etal-2018-pre} with the same set of 20 languages as \citet{philip-etal-2020-monolingual}.\footnote{\{en, ar, he, ru, ko, it, ja, zh\_cn, es, fr, pt\_br, nl, tr, ro, pl, bg, vi, de, fa, hu\}} This corpus is multi-parallel, i.e., it has training data for all 380 (20$\times$19) language pairs (see Table~\ref{tab:ted_training_data} in Appendix for detailed statistics). It also includes official valid and test splits for all these language pairs.

We train the English-centric models for 120 epochs ($\approx$1.8M updates). The Base 12-2 English-centric model is initialized from Base 6-6 at epoch 60 and trained for another 60 epochs, using the procedure described in Section~\ref{sec:shallow_dec}.
These models are then fine-tuned with multi-parallel data for another 10 epochs ($\approx$1.4M updates)\footnote{Note that an ``epoch'' when using multi-parallel data corresponds to approximately 9 English-centric epochs in terms of updates.} into single-decoder Transformers or Hybrid models or multi-decoder Transformers. We create a shared BPE model with 64k merge operations (vocabulary size 70k) and with inline casing \cite{berard-etal-2019-naver}. More hyper-parameters are given in Appendix~\ref{appendix:hyperparameters}.

\subsection{Evaluation settings}
\label{sec:ted_evaluation_settings}
The TED Talks models are evaluated on the provided multi-parallel validation and test sets. Since those are already word-tokenized, we run SacreBLEU with the \texttt{-{}-tok none} option.\footnote{SacreBLEU signature:\\ \texttt{BLEU+c.mixed+\#.1+s.exp+tok.none+v.1.5.1}}

We report average test BLEU scores into English ($\to$EN, 19 directions), from English ($\leftarrow$EN, 19 directions) and outside of English (/~EN, 342 directions).
We also compute the decoding speed in Words Per Second (WPS)\footnote{Not BPE tokens per second, as we do not want the speed measurement to depend on the BPE tokenization used.} when translating the concatenated $\to$EN valid sets on a V100 with batch size 64 and beam size 5 (averages over 3 runs). Additional speed benchmarks with other decoding settings and time spent in each component are given in Appendix Table~\ref{tab:ted_speed_benchmark}. We also report chrF scores and results on more models in Appendix Table~\ref{tab:ted_scores_more}.

\subsection{Position of the language code}
The prevalent approach in multilingual NMT for choosing the target language is to prefix the source sequence with a \emph{language code} \cite{johnson-etal-2017-googles}. However, it is also possible, like \citet{tang2020multilingual}, to put this code on the target side. Table~\ref{tab:ted_lang_codes} in Appendix analyzes the impact of the position of this language code on BLEU performance. Like observed by \citet{wu-etal-2021-language}, decoder-side language codes result in very low zero-shot performance in the English-centric setting. They also degrade the performance of the Base 12-2 models in all translation directions. For this reason, all our experiments use source-side language codes.

\subsection{BLEU results}
Table~\ref{tab:ted_scores} evaluates the techniques we proposed in Section~\ref{sec:techniques} on TED Talks. First, we see that the Base 12-2 models (\modelref{ted_base_12_2}, \modelref{ted_base_12_2_multipara}) perform as well or better as the Base 6-6 models (\modelref{ted_base_6_6}, \modelref{ted_base_6_6_multipara}) in all language directions, with a $1.7\times$ speed boost. Multi-parallel fine-tuning (\modelref{ted_base_6_6_multipara}, \modelref{ted_base_12_2_multipara}) significantly increases translation quality between non-English languages and incurs no drop in performance in the $\leftrightarrow$EN directions.
Test-time filtering of the vocabulary with $K_{test}=10$ (see Section~\ref{sec:vocab_filtering}) does not degrade BLEU but increases decoding speed by 30\% (\modelref{ted_base_12_2_multipara_test_filter}). More aggressive filtering with $N_{test}=4$k results in a drop in BLEU (\modelref{ted_base_12_2_multipara_test_filter_more}).\footnote{Note that when $N=4$k, we also apply a frequency threshold of $K=10$ on BPE tokens and characters.} The latter leads to slightly longer outputs (in terms of BPE units), which explains why it is not faster. When training with $N_{train}=4$k, we can get the same speed boost (\modelref{ted_base_6_6_multipara_train_filter}, \modelref{ted_base_12_2_multipara_train_filter}), without any drop in BLEU compared to models without BPE filtering (\modelref{ted_base_6_6_multipara}, \modelref{ted_base_12_2_multipara}).

Then, we observe that Hybrid models (\modelref{ted_hybrid_12_2}, \modelref{ted_hybrid_12_3}) are slightly worse than Transformers in terms of BLEU, but are also much faster at decoding. Hybrid 12-2 (\modelref{ted_hybrid_12_2}) is $3\times$ faster than Base 6-6 (with -0.2 BLEU on average) and $1.7\times$ faster than Base 12-2 (with -0.3 BLEU). Hybrid 12-3 (\modelref{ted_hybrid_12_3}) is slower than Hybrid 12-2 and not clearly better in terms of BLEU (+0.1 BLEU).

Finally, we see that fine-tuning the English-centric 12-2 model into 20 language-specific shallow decoders with the multi-parallel data (\modelref{ted_base_12_2_multidec}) results in the highest BLEU scores overall, with the same speed benefits as with a single shallow decoder (\modelref{ted_base_12_2_multipara_train_filter}). A Base 6-6 model can also be fine-tuned into multiple 2-layer language-specific decoders (\modelref{ted_base_6_2_multidec}) and get the same performance as the single Base 6-6 or Base 12-2 models (\modelref{ted_base_6_6_multipara_train_filter}, \modelref{ted_base_12_2_multipara_train_filter}). This is convenient if one wants to quickly improve the decoding speed of existing 6-6 models.

Lastly, we do a similar set of experiments within a different framework and observe the same trends (see Table~\ref{tab:reproducibility} in Appendix). 


\begin{table}[t]
	\hspace{-.45cm}
	\begin{tabular}{c@{\hspace{.2cm}}c|c|c|c|c}
		& Model & $\to$EN & $\leftarrow$EN & $/$ EN & WPS \\
		\cline{2-6}
		& \multicolumn{5}{c}{SOTA \citep{philip-etal-2020-monolingual}} \\
		\cline{2-6}
		& Bilingual & 32.4 & 24.4 & 15.0 & -- \\   
		& Best multi & 32.3 & 24.1 & 15.8 & -- \\        
		\cline{2-6}
		& \multicolumn{5}{c}{English-centric} \\
		\cline{2-6}
		\labelledmodelcounter{ted_small_6_6} & Small 6-6 & 31.6 & 23.1 & 11.6 & 703 \\ 
		\labelledmodelcounter{ted_base_6_6} & Base 6-6 & 31.8 & 24.2 & 13.5 & 753 \\ 
		\labelledmodelcounter{ted_base_12_2} & Base 12-2 & 33.6 & 24.3 & 14.1 & 1287 \\
		\labelledmodelcounter{ted_base_12_2_pivot} & + EN pivot & -- & -- & 15.4 & -- \\
		\cline{2-6}
		& \multicolumn{5}{c}{+ Multi-parallel} \\
		\cline{2-6}
		\labelledmodelcounter{ted_base_6_6_multipara} & Base 6-6 & 32.8 & 24.3 & 16.3 & 732 \\ 
		\labelledmodelcounter{ted_base_12_2_multipara} & Base 12-2 & 33.5 & 24.5 & 16.3 & 1203 \\ 
		\labelledmodelcounter{ted_base_12_2_multipara_test_filter} & + $K_{test}=10$ & 33.4 & 24.3 & 16.2 & 1539 \\ 
		\labelledmodelcounter{ted_base_12_2_multipara_test_filter_more} & + $N_{test}=4$k & 31.5 & 22.5 & 15.2 & 1457 \\ 
		\cline{2-6}
		& \multicolumn{5}{c}{+ BPE filtering ($N_{train}=4$k)} \\
		\cline{2-6}
		\labelledmodelcounter{ted_base_6_6_multipara_train_filter} & Base 6-6 & 32.9 & 24.2 & 16.3 & 789 \\ 
		\labelledmodelcounter{ted_base_12_2_multipara_train_filter} & Base 12-2 & 33.3 & 24.3 & 16.3 & 1552 \\ 
		\labelledmodelcounter{ted_hybrid_12_2} & Hybrid 12-2 & 32.8 & 23.5 & 16.1 & \textbf{2422} \\ 
		\labelledmodelcounter{ted_hybrid_12_3} & Hybrid 12-3 & 32.9 & 23.7 & 16.1 & 2145 \\ 
		\cline{2-6}
        & \multicolumn{5}{c}{+ Multi-decoder} \\
        \cline{2-6}
        \labelledmodelcounter{ted_base_6_2_multidec} & 6-6 $\to$ 6-2 & 33.0 & 24.2 & 16.0 & 1608 \\ 
		\labelledmodelcounter{ted_base_12_2_multidec} & 12-2 $\to$ 12-2 & \textbf{33.8} & \textbf{25.1} & \textbf{16.7} & 1614 \\ 
	\end{tabular}
	\caption{Test BLEU scores and decoding speed of \textbf{TED Talks models} of various depths. SOTA's ``best multi'' is a Transformer Small 6-6 multi-parallel model with adapter layers. WPS: speed in words per second for $\to$EN translation. Table~\ref{tab:ted_scores_more} in Appendix reports scores by more models and with additional metrics.}
	\label{tab:ted_scores}
\end{table}

\section{ParaCrawl Experiments}

\subsection{Data and hyper-parameters}
We scale our experiments to a more realistic setting, with the same number of languages as before, but larger amounts of training data and larger models.

We download ParaCrawl v7.1 \cite{banon-etal-2020-paracrawl} in the 19 highest-resource languages paired with English.\footnote{\{fr, de, es, it, pt, nl, nb, cs, pl, sv, da, el, fi, hr, hu, bg, ro, sk, lt\}} Then, like \citet{freitag-firat-2020-complete}, we build a multi-parallel corpus by aligning all pairs of languages through their English side. See Table~\ref{tab:paracrawl_training_data} in Appendix for training data statistics. We train a shared BPE model with 64k merge operations and inline casing by sampling from this data with temperature 5 (final vocabulary size: 69k).

We train the English-centric models for 1M steps and fine-tune them with multi-parallel data for 200k more steps. Hybrid and Multi-decoder models are also fine-tuned for 200k steps from the English-centric models with multi-parallel data. Big 6-6 bilingual baselines are trained with the same hyper-parameters for 120k steps, with joint BPE vocabularies of size 16k.  More hyper-parameters are given in Appendix~\ref{appendix:hyperparameters}.

The Big 6-6 and Big 12-2 English-centric models took each around 17 days to train on 4 A100s. The multi-parallel fine-tuning stages (single/multi-decoder and hybrid) took $\approx$4.5 days on 2 A100s each.

\subsection{Evaluation settings}
The ParaCrawl models are evaluated on our own valid and test splits from TED2020 \cite{reimers-gurevych-2020-making}.\footnote{TED2020 is a different crawl from TED than that of the ``TED Talks'' corpus. It is more recent, it has more data and languages and it is not word-tokenized.} We shuffle the parallel corpus for each translation direction and take 3000 line pairs for the validation set and 3000 for the test set. To compare against the state of the art, we also provide scores on standard test sets from WMT for some language pairs. In both cases, we use SacreBLEU with its default options.\footnote{SacreBLEU signature:\\ \texttt{BLEU+c.mixed+\#.1+s.exp+tok.13a+v.1.5.1}}

Like in Section~\ref{sec:ted}, we compute average $\to$EN, $\leftarrow$EN and /~EN test BLEU scores and WPS on $\to$EN TED2020 valid. Table~\ref{tab:paracrawl_scores_more} in Appendix reports TED2020 test chrF \cite{popovic-2015-chrf}, as well as spBLEU scores on FLORES devtest \cite{goyal2021flores101}.

\subsection{BLEU results}

\begin{table}[t]
    \hspace{-.5cm}
	\begin{tabular}{c@{\hspace{.2cm}}c|c|c|c|c}
		& Model & $\to$EN & $\leftarrow$EN & $/$ EN & WPS \\
		\cline{2-6}
		& \multicolumn{5}{c}{English-centric} \\
		\cline{2-6}
		\labelledmodelcounter{para_base_6_6} & Base 6-6 & 31.4 & 26.0 & 13.2 & 656 \\ 
		\labelledmodelcounter{para_big_6_6} & Big 6-6 & 35.0 & 29.0 & 14.4 & 623 \\ 
		\labelledmodelcounter{para_big_12_2} & Big 12-2 & \textbf{35.5} & \textbf{29.6} & 16.7 & 1033 \\ 
		\labelledmodelcounter{para_big_12_2_pivot} & + EN pivot & -- & -- & 21.5 & -- \\
		\cline{2-6}
		& \multicolumn{5}{c}{+ Multi-parallel} \\
		\cline{2-6}
		\labelledmodelcounter{para_big_6_6_multipara} & Big 6-6 & 34.2 & 28.3 & 20.7 & 595 \\ 
		\labelledmodelcounter{para_big_12_2_multipara} & Big 12-2 & 34.9 & 29.2 & 21.3 & 1030 \\ 
		\labelledmodelcounter{para_big_12_2_multipara_test_filter} & + $N_{test}=16$k & 34.8 & 29.0 & 21.2 & 1328 \\ 
		\labelledmodelcounter{para_big_12_2_multipara_test_filter_more} & + $N_{test}=8$k & 32.9 & 27.8 & 20.4 & 1283 \\ 
		\cline{2-6}
		& \multicolumn{5}{c}{+ BPE filtering ($N_{train}=8$k)} \\
		\cline{2-6}
		\labelledmodelcounter{para_big_6_6_multipara_train_filter} & Big 6-6 & 34.0 & 28.4 & 20.7 & 679 \\ 
		\labelledmodelcounter{para_big_12_2_multipara_train_filter} & Big 12-2 & 34.8 & 29.1 & 21.2 & 1261 \\ 
		\labelledmodelcounter{para_hybrid_12_2} & Hybrid 12-2 & 33.9 & 28.3 & 20.7 & \textbf{1796} \\ 
 		\labelledmodelcounter{para_hybrid_12_3} & Hybrid 12-3 & 34.0 & 28.3 & 20.8 & 1659 \\ 
		\cline{2-6}
		& \multicolumn{5}{c}{+ Multi-decoder} \\
		\cline{2-6}
		\labelledmodelcounter{para_big_6_2_multidec} & 6-6 $\to$ 6-2 & 34.0 & 28.8 & 21.0 & 1333 \\ 
		\labelledmodelcounter{para_big_12_2_multidec} & 12-2 $\to$ 12-2 & 35.3 & 29.4 & \textbf{21.8} & 1270 \\ 
	\end{tabular}
	\caption{TED2020 test BLEU scores and decoding speed of \textbf{ParaCrawl models} of various depths. WPS: speed in words per second for $\to$EN translation. Table~\ref{tab:paracrawl_scores_more} in Appendix reports scores by more models and with additional metrics.}
	\label{tab:paracrawl_scores}
\end{table}

Table~\ref{tab:paracrawl_scores} shows that, like in the TED Talks experiments, the 12-2 architecture (\modelref{para_big_12_2}, \modelref{para_big_12_2_multipara}) gets as good or better BLEU scores than the standard 6-6 Transformer (\modelref{para_big_6_6}, \modelref{para_big_6_6_multipara}) and is 70\% faster. It outperforms the Big 6-6 baseline in all 38 English-centric directions, both according to BLEU and chrF.
Test-time BPE filtering with $N_{test}=16$k (see Section~\ref{sec:vocab_filtering}) does not degrade BLEU (\modelref{para_big_12_2_multipara_test_filter}) and improves decoding speed by 30\%. However, $N_{test}=8$k leads to a large drop in BLEU (\modelref{para_big_12_2_multipara_test_filter_more}) without any adddional speed benefit.\footnote{Note that when $N_{train}$ or $N_{test}$ is set, we additionally apply a frequency threshold of $K=100$ on BPE tokens and characters.} Indeed, in this setting 1.5\% of the tokens that would have been generated by the non-filtered model become out-of-vocabulary.\footnote{Average number on the $\leftarrow$EN TED2020 valid outputs of the Big 12-2 multi-parallel model.} This means that the filtered model has to settle for tokens that are possibly further from the true data distribution, accentuating the exposure bias (and possibly leading to degenerate outputs). Like with TED Talks, this issue is solved when training with BPE filtering (\modelref{para_big_12_2_multipara_train_filter}). $N_{train}=8$k leads to vocabularies of size \numprint{8405} on average at the cost of 4.2\% longer target sequences.\footnote{Average number on the training data.} The multi-parallel Big 12-2 model with train-time BPE filtering (\modelref{para_big_12_2_multipara_train_filter}) also performs better than its Big 6-6 counterpart (\modelref{para_big_12_2_multipara_train_filter}) and is almost twice faster. It outperforms the latter in 370 out of 380 translation directions according to BLEU, and in 377 directions according to chrF. It also gets the same $\leftrightarrow$EN performance as the English-centric Big 6-6 model (\modelref{para_big_6_6}).
Interestingly, pivot translation with an English-centric model is a strong baseline on /~EN (\modelref{para_big_12_2_pivot}), slightly better than direct translation with the models fine-tuned on multi-parallel data (but also twice slower).
Like on TED Talks, the Hybrid 12-2 model (\modelref{para_hybrid_12_2}) provides a very good BLEU/speed tradeoff, matching the quality of a similar Transformer Big 6-6 model (\modelref{para_big_6_6_multipara_train_filter}) at $2.6\times$ the speed. The Big 12-2 multi-decoder model (\modelref{para_big_12_2_multidec}) slightly outperforms the single-decoder model in all directions (\modelref{para_big_12_2_multipara_train_filter}), matching the $\leftrightarrow$EN performance of the best English-centric model.

\begin{table}
	\hspace{-.2cm}
	\begin{tabular}{c|c|c|c}
		Model & DE-EN & EN-DE & DE-FR \\
		\hline
        \citet{shi-etal-2020-oppos} & 40.7 & \textbf{42.6} & \textbf{35.4} \\
		\hline
		Bilingual Big 6-6 & \textbf{42.3} & 42.0 & 34.9 \\
        Mul. Big 6-6 (\modelref{para_big_6_6_multipara_train_filter}) & 38.1 & 36.7 & 32.4 \\
		Mul. Big 12-2 (\modelref{para_big_12_2_multipara_train_filter}) & 38.4 & 37.2 & 32.6 \\
	\end{tabular}
	\caption{Comparison of our multi-parallel ($N_{train}=8$k) \textbf{ParaCrawl models} with bilingual baselines and with the state of the art on \emph{newstest2019}. \citet{shi-etal-2020-oppos} is the top-ranking submission at WMT20 in those languages. We report their baseline scores (i.e., without back-translation, ensembling, etc.) Pivot translation with bilingual models (resp. with Big 12-2 English-centric) gives 34.2 BLEU (resp. 32.0 BLEU) on DE-FR.}
	\label{tab:paracrawl_sota}
\end{table}

\begin{figure}
    \centering
    \includegraphics[width=0.5\textwidth]{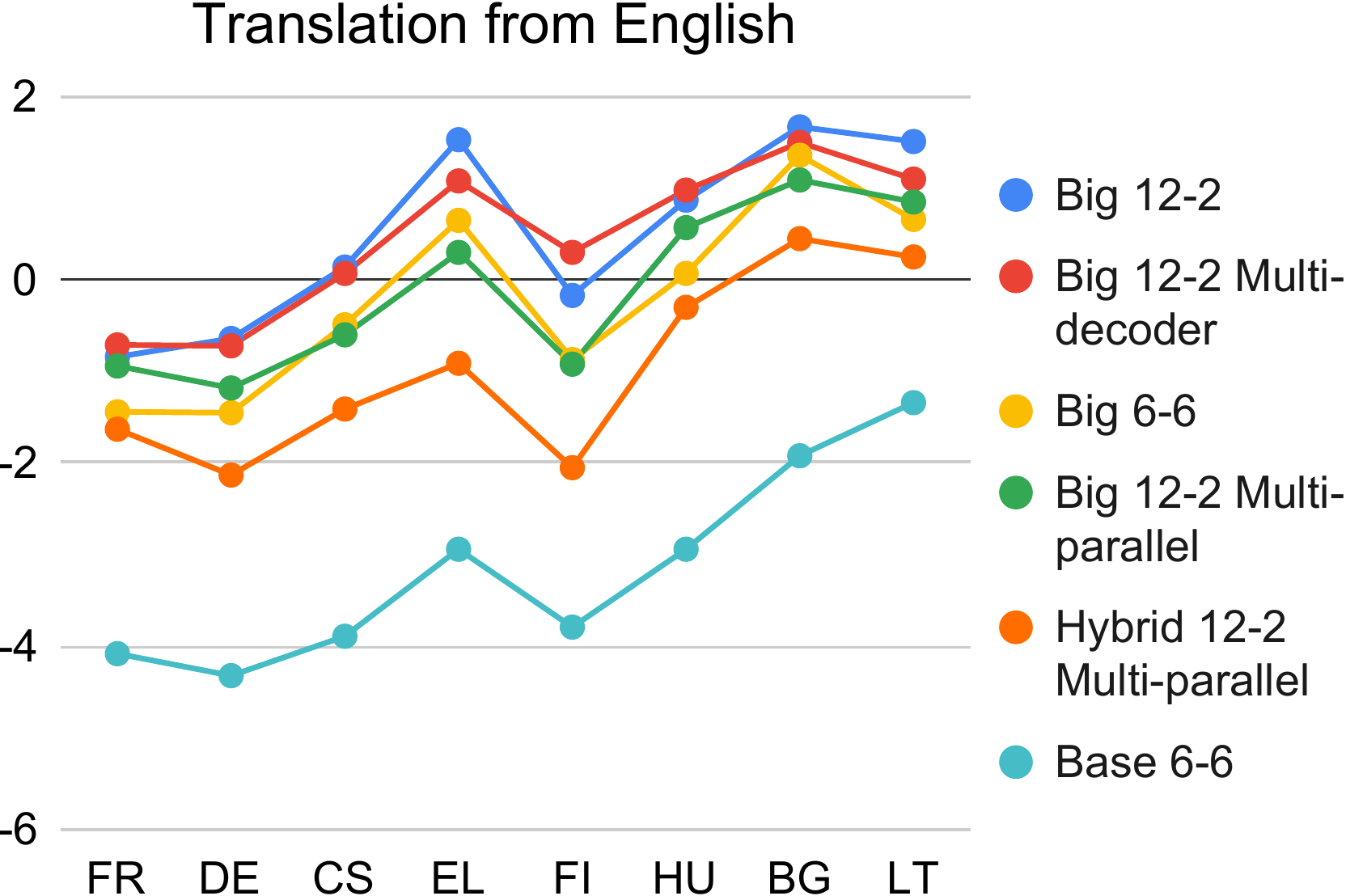}
    \caption{BLEU deltas on $\leftarrow$EN TED2020 test (in a subset of 8 languages) by our multilingual \textbf{ParaCrawl models} (\modelref{para_big_12_2}, \modelref{para_big_12_2_multidec}, \modelref{para_big_6_6}, \modelref{para_big_12_2_multipara_train_filter}, \modelref{para_hybrid_12_2}, \modelref{para_base_6_6}) compared to bilingual Big 6-6 baselines. The languages are sorted from highest-resource to lowest-resource.}
    \label{fig:paracrawl_bleu_delta}
\end{figure}

Table~\ref{tab:paracrawl_sota} compares our multi-parallel models with bilingual Big 6-6 baselines and with reported numbers in the literature. It shows that bilingual models trained on ParaCrawl-only can reach similar performance as well-trained WMT baselines.

Figure~\ref{fig:paracrawl_bleu_delta} shows the $\leftarrow$EN BLEU difference between our multilingual models and the ParaCrawl bilingual baselines on a subset of 8 languages. We see the same trend as in the literature: multilingual training hurts performance on high-resource languages and helps on lower-resource languages. We also see that Transformer Big 12-2 consistently outperforms Big 6-6 and that multi-parallel training consistently hurts $\leftarrow$EN performance. Figure~\ref{fig:paracrawl_bleu_delta_non_english} in Appendix shows similar scores for the $\to$EN and /~EN directions

\subsection{Impact of framework}
Recent work by \citet{narang2021transformer} suggest that the implementation framework can change the conclusions one makes about Transformer-based architectures. In addition to a PyTorch-based framework (fairseq, \citealp{ott-etal-2019-fairseq}), we conduct TED Talks experiments with an in-house TensorFlow implementation, whose results are shown in Appendix (Table~\ref{tab:reproducibility}). Although BLEU and WPS values are a bit different, we observe the same trends. This confirms that our TED Talks experiments can be reproduced in a completely different framework with the same observations. 

\subsection{Impact of sequence length}

\begin{figure}
    \includegraphics[width=0.5\textwidth]{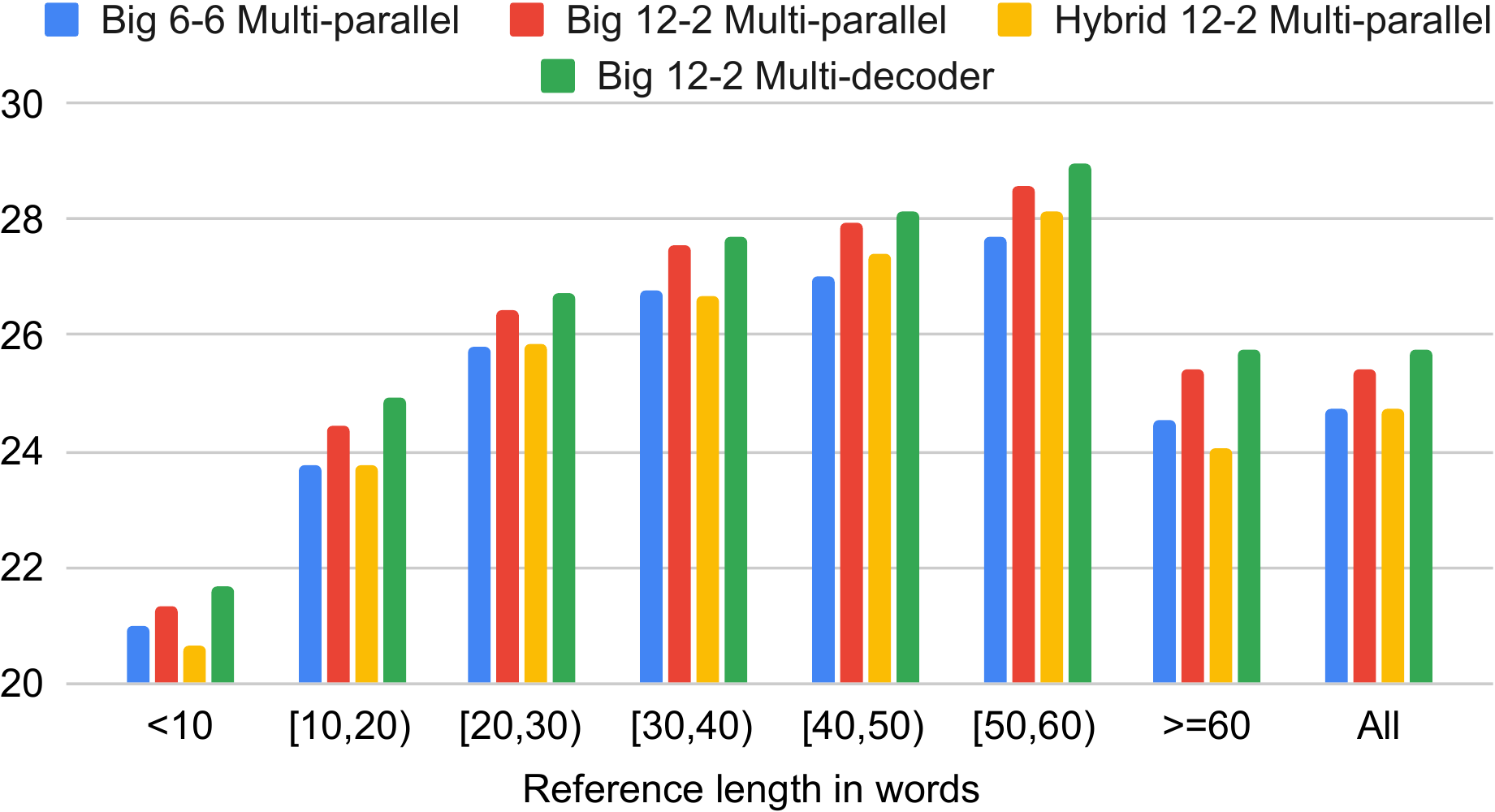}
    \caption{BLEU scores on $\leftarrow$EN TED2020 test by the \textbf{ParaCrawl models} (\modelref{para_big_6_6_multipara_train_filter}, \modelref{para_big_12_2_multipara_train_filter}, \modelref{para_hybrid_12_2}, \modelref{para_big_12_2_multidec}) according to sentence length.}
    \label{fig:bleu_by_length_from_en}
\end{figure}

When reducing the depth of the decoder, one could expect that it would have trouble generating long sequences. Figure~\ref{fig:bleu_by_length_from_en} reports BLEU scores for different length buckets. We observe no abnormal patterns in any of the proposed architectures. We first note that Big 12-2 (\modelref{para_big_12_2_multipara_train_filter}) performs consistently better than Big 6-6 (\modelref{para_big_6_6_multipara_train_filter}) across all sentence lengths. The performance of the Hybrid 12-2 model (\modelref{para_hybrid_12_2}) is also consistent (slightly lower than Transformers). Figure~\ref{fig:bleu_by_length_more} in Appendix shows scores by length in the $\to$EN direction and with greedy decoding.

\subsection{Robustness analysis}
Even if different decoder architectures reach similar BLEU performance, some architectures might be more brittle to noise than others. To test each model's robustness, we introduce synthetic noise by either adding an unknown character (\textit{unk}) randomly at the beginning, middle, or end of the sentence; or by applying 3 random char-level operations (del, ins, swap, or sub) (\textit{char}). Table \ref{tab:robustness} reports the BLEU consistency (``Cy BLEU'') as introduced by \citet{niu2020} on $\leftarrow$EN translation.\footnote{BLEU consistency measures the similarity between the translations by the same model of the clean sentence and its noised version.} As previously, deep-encoder / shallow decoder models (Big 12-2, Big 12-2 Multi-decoder) outperform the other architectures.
BPE filtering slightly hurts robustness, despite showing close BLEU scores on the clean test sets.
Additional results are given in the Appendix (Table~\ref{tab:appendix:robustness}).  

\begin{table}[]
    \begin{tabular}{c@{\hspace{.2cm}}l|c|c}
        & Model & \thead{Cy BLEU \\ unk }  & \thead{Cy BLEU \\ char }  \\
        \hline
        \modelref{para_big_6_6_multipara} & Big 6-6 & 73.3 & 54.2 \\
        \modelref{para_big_12_2_multipara} & Big 12-2 & \textbf{76.4} & \textbf{56.1}\\
        \modelref{para_big_12_2_multipara_test_filter} & Big 12-2 ($N_{test}=16$k) & 75.9    &56.0 \\
        \modelref{para_big_6_6_multipara_train_filter} & Big 6-6 ($N_{train}=8$k) & 74.4 & 54.5  \\
        \modelref{para_big_12_2_multipara_train_filter} & Big 12-2 ($N_{train}=8$k) & 73.7    &55.5 \\
        \modelref{para_hybrid_12_2} & Hybrid 12-2 & 75.0 & 55.3  \\ 
        \modelref{para_big_12_2_multidec} & Big 12-2 Multi-decoder & 76.1  & 55.4      \\
    \end{tabular}
    \caption{Robustness evaluation with average BLEU consistency \cite{niu2020} on $\leftarrow$EN test sets by the multi-parallel \textbf{ParaCrawl models}.}
    \label{tab:robustness}
\end{table}

\subsection{Human evaluation}

We conduct a human evaluation to compare the English-centric Big 6-6 and Big 12-2 models. It is done by certified professionals who are proficient in both the source and target language. We use bilingual direct assessment (DA), where raters have to evaluate the adequacy and fluency of each translation on a 0-5 scale given the source sentence. We select a random subset of 200 sentences from \emph{newstest2019} for DE-EN / EN-DE and from \emph{newstest2014} for FR-EN / EN-FR.\footnote{We ensure that the source sentences are original text in the corresponding language to avoid biased evaluation results due to Translationese.} For each translation direction, 3 raters are shown all the source sentences and their translations by both systems in random order.  Table \ref{tab:humaneval} reports relative results averaged across the 3 raters. Big 12-2 outperforms Big 6-6 in 3 out of 4 language directions. 
Contrary to \citet{kong-etal-2021-multilingual} and according to both human evaluation and automatic metrics, our single-shallow-decoder model performs at least as well as the baseline model.\footnote{Note that our human evaluation results are only on high-resource languages, but \citet{kong-etal-2021-multilingual} observed the largest BLEU drop on high-resource languages.}

\begin{table}
	\centering
	\begin{tabular}{c|c|c|c}
		Model & 12-2 > 6-6 & 12-2 = 6-6 & 12-2 < 6-6 \\
		\hline
		EN$\to$FR & \textbf{26\%} & 51\% & 23\% \\
		FR$\to$EN & \textbf{25\%} & 51\% & 24\% \\
		EN$\to$DE & \textbf{31\%} & 44\% & 25\% \\
		DE$\to$EN & 24\% & 50\% & \textbf{26\%} \\
	\end{tabular}
	\caption{Human evaluation on English-centric \textbf{ParaCrawl models} (\modelref{para_big_6_6} and \modelref{para_big_12_2}).}
	\label{tab:humaneval}
\end{table}

\section{Conclusion}
On one hand, multilingual NMT saves training and deployment costs. On the other hand, larger architectures (required to keep performance on a par with bilingual MT) and large shared vocabularies penalize inference speed and user latency. In this work, we study various approaches to improve the speed of multilingual models without degrading translation quality. We find that Transformers with a deep encoder and a shallow decoder can outperform a baseline Transformer at a much faster decoding speed. This can be combined with per-language vocabulary filtering to reach a global $2\times$ speed-up with no loss in BLEU. A careful analysis of the results on different aspects such as sequence length, robustness to noise, and human evaluation validates this finding. Additionally, language-specific shallow decoders can be trained to get even better performance at the same speed. And finally, hybrid models with a shallow RNN decoder offer an excellent BLEU-speed trade-off ($3\times$ faster than baseline with a minor drop in BLEU). We also provide supplementary material (code, checkpoints, test splits, translation outputs) to facilitate reproducibility.\footnote{\url{https://europe.naverlabs.com/research/natural-language-processing/efficient-multilingual-machine-translation}}


\bibliography{main}
\bibliographystyle{acl_natbib}

\appendix
\onecolumn

\section{Appendix}

\subsection{Position of the language code}

Table~\ref{tab:ted_lang_codes} analyzes the impact of the language code position on BLEU performance.
With the Base 6-6 architecture, decoder-side codes perform approximately as well as encoder-side codes (except for zero-shot translation). However, with the Base 12-2 architecture, decoder-side codes result in a noticeable drop in performance in most directions.
Indeed, when the lang code is on the source side, the deep encoder knows the target language and can start ``translating.'' When it is on the target side, the encoder has no way of knowing which language to start translating into. So it outputs a universal representation that is harder to transform into a target-language sentence by the limited-capacity shallow decoder.
Note that $\to$EN performance in the English-centric setting is not affected. We believe this is because the encoder can easily guess that the target language is English by detecting the language of the input. We believe this is also the reason for the low zero-shot performance: the encoder starts translating all non-English inputs into English, and the decoder receives a representation that it cannot translate into other languages than English.

\begin{table}[h]
	\centering
	\begin{tabular}{c@{\hspace{.2cm}}c|c|c|c|c}
		& Lang code position & \multicolumn{1}{c|}{Model} & $\to$EN & $\leftarrow$EN & $/$ EN \\
		\hline
		& \multicolumn{5}{c}{English-centric} \\
		\hline
		\modelref{ted_base_6_6} & Encoder & Base 6-6 & 31.8 & 24.2 & 13.5 \\ 
		\labelledmodelcounter{ted_base_6_6_lang_as_bos} & Decoder & Base 6-6 & 32.0 & 23.9 & 5.43 \\ 
		\modelref{ted_base_12_2} & Encoder & Base 12-2 & \textbf{33.6} & \textbf{24.3} & \textbf{14.1} \\ 
		\labelledmodelcounter{ted_base_12_2_lang_as_bos} & Decoder & Base 12-2 & 33.5 & 22.8 & 0.74 \\ 
		\hline
		& \multicolumn{5}{c}{Multi-parallel + BPE filtering ($N_{train}=4$k)} \\
		\hline
		\modelref{ted_base_6_6_multipara_train_filter} & Encoder & Base 6-6 & 32.9 & 24.2 & 16.3 \\ 
		\labelledmodelcounter{ted_base_6_6_multipara_train_filter_lang_as_bos} & Decoder & Base 6-6 & 32.7 & 23.9 & 16.2 \\ 
		\modelref{ted_base_12_2_multipara_train_filter} & Encoder & Base 12-2 & \textbf{33.3} & \textbf{24.3} & \textbf{16.3} \\ 
		\labelledmodelcounter{ted_base_12_2_multipara_train_filter_lang_as_bos} & Decoder & Base 12-2 & 32.5 & 23.0 & 15.7 \\ 
	\end{tabular}
	\caption{Test BLEU scores of \textbf{TED Talks models} with encoder-side or decoder-side language codes. Like \citet{tang2020multilingual}, we implement decoder-side lang codes by replacing \texttt{BOS} (i.e., the first embedding input to the decoder) with the lang code.}  
	\label{tab:ted_lang_codes}
\end{table}

\subsection{Framework and hyper-parameters}
\label{appendix:hyperparameters}

We do our experiments in the fairseq v0.10.2 framework \cite{ott-etal-2019-fairseq}, which we modify to implement on-the-fly pre-processing and sampling from multilingual corpora.

We randomly sample language pairs with $p_k=\frac{D_k^{1/T}}{\sum{D_i^{1/T}}}$ where $D_k$ is the number of sentence pairs for language pair $k$ and $T$ is the temperature \cite{arivazhagan2019massively}. Tables~\ref{tab:ted_training_data} and~\ref{tab:paracrawl_training_data} give the resulting sampling probabilities by target language. We build heterogeneous batches using this sampling strategy (i.e., containing any mixture of languages), by sampling 100k sentence pairs at a time and sorting them by length into batches. Language-specific decoders are trained with homogeneous batches with respect to the target language (we increase the ``buffer size'' to 1M and group sentence pairs by target language before batching).

Tables~\ref{tab:ted_hyperparameters} and~\ref{tab:paracrawl_hyperparameters} give the fairseq hyperparameters of our \textbf{TED Talks} and \textbf{ParaCrawl} Transformer models. Tables~\ref{tab:ted_training_details} and~\ref{tab:paracrawl_training_details} give the training details of the fine-tuned models.

Our Hybrid models use a variant of the hybrid RNMT+ architecture proposed by \citet{chen2018best}. Contrary to them, we use single-head additive attention \cite{bahdanau2014neural}; \textit{sum} the attention and LSTM output before the vocabulary projection; and apply layer normalization on the input of the LSTMs (rather than on the gates). We apply the same amounts of dropout as in the Transformer but on both the LSTM outputs (except for the first LSTM) and the target embeddings. 

\newpage

\begin{table}[h]
	\centering
	\begin{tabular}{c|c|c|c}
Language & Code & English-centric lines & Multi-parallel lines \\
\hline
English & en & 3,556,962 & 3,556,962 \\
Arabic & ar & 214,111 & 3,430,385 \\
Hebrew & he & 211,819 & 3,399,679 \\
Russian & ru & 208,458 & 3,379,440 \\
Korean & ko & 205,640 & 3,350,599 \\
Italian & it & 204,503 & 3,350,483 \\
Japanese & ja & 204,090 & 3,312,997 \\
Mandarin Chinese & zh\_cn & 199,855 & 3,297,628 \\
Spanish & es & 196,026 & 3,234,798 \\
French & fr & 192,304 & 3,192,551 \\
Brazilian Portuguese & pt\_br & 184,755 & 3,110,048 \\
Dutch & nl & 183,767 & 3,053,593 \\
Turkish & tr & 182,470 & 3,017,706 \\
Romanian & ro & 180,484 & 3,055,943 \\
Polish & pl & 176,169 & 3,002,206 \\
Bulgarian & bg & 174,444 & 2,946,693 \\
Vietnamese & vi & 171,995 & 2,807,695 \\
German & de & 167,888 & 2,900,115 \\
Persian & fa & 150,965 & 2,405,646 \\
Hungarian & hu & 147,219 & 2,465,081 \\
\hline
Total & all & 7,113,924 & 62,270,248 \\
	\end{tabular}
	\vspace{-.1cm}
	\caption{Size of the \textbf{Top 20 \href{https://github.com/neulab/word-embeddings-for-nmt}{TED Talks}} corpus. English has 253,292 unique lines. The average English sentence length is 21.1 words and 26.7 wordpieces.}
	\label{tab:ted_training_data}
\end{table}

\begin{table}[h]
	\centering
	\begin{tabular}{c@{\hspace{.2cm}}c|c|c|c}
		& Model & $\to$EN & $\leftarrow$EN & $/$ EN \\
		\hline
		& \multicolumn{4}{c}{English-centric} \\
		\hline
		\modelref{ted_base_6_6} & Base 6-6 post-norm & 32.1 & \textbf{24.5} & 13.6 \\
		\modelref{ted_base_12_2} & Base 12-2 post-norm pre-trained & \textbf{33.7} & \textbf{24.5} & 14.1 \\
		\hline
		\labelledmodelcounter{ted_base_12_2_prenorm} & Base 12-2 pre-norm & 33.5 & 24.3 & 12.6 \\
		\labelledmodelcounter{ted_base_12_2_enc_prenorm} & Base 12-2 enc pre-norm & 33.3 & 24.1 & 13.1 \\
		\labelledmodelcounter{ted_base_18_1_enc_prenorm} & Base 18-1 enc pre-norm$^\star$ & 31.6 & 22.8 & 12.1 \\  
	\end{tabular}
	\caption{Valid BLEU scores of English-centric \textbf{TED Talks models} with deep encoders, depending on the training strategy used. $\star$: this model was stopped before the end (after 60 epochs).}
	\label{tab:deep_training_strategy}
\end{table}

\begin{figure}[H]
    \centering
	\includegraphics[width=0.45\textwidth]{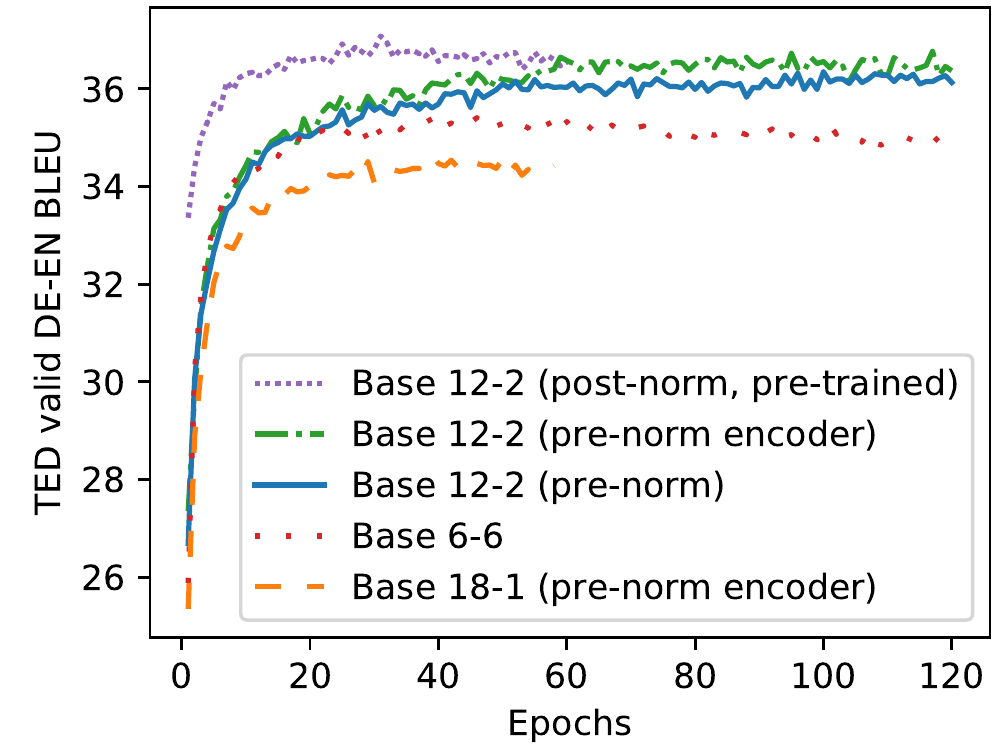}
	\includegraphics[width=0.45\textwidth]{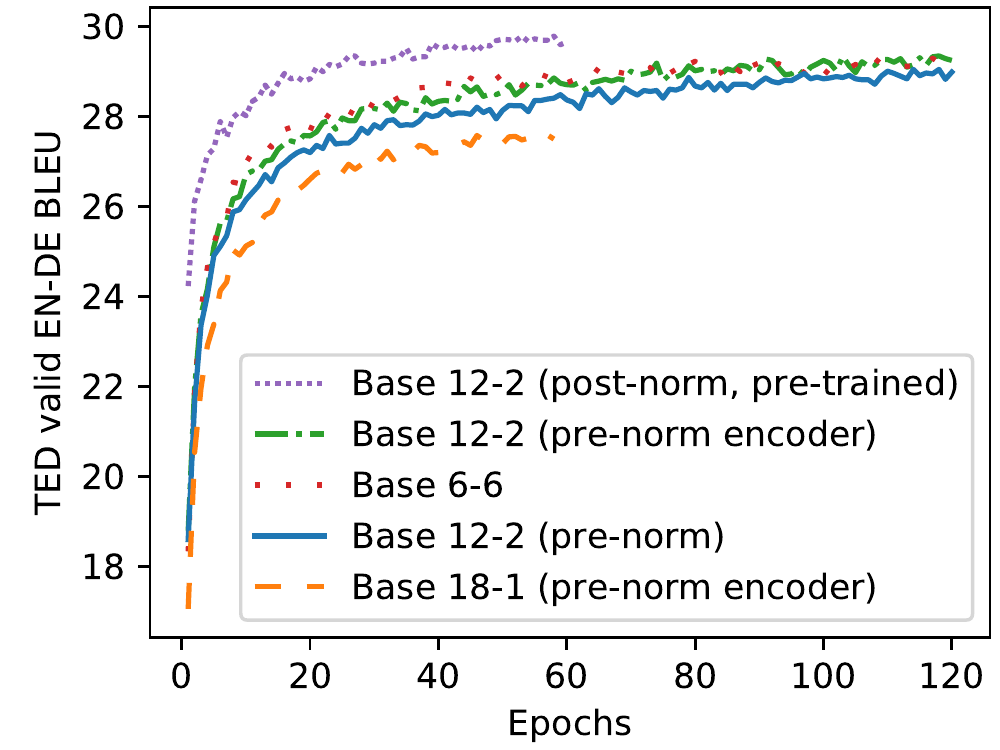}
	\caption{Training progress of English-centric models trained on \textbf{TED Talks}. The names in the legend are sorted from highest to lowest BLEU score. The pre-trained 12-2 model was initialized with the 6-6 model's checkpoint at epoch 60 and fine-tuned for 60 more epochs. The 18-1 model was stopped in the middle of training.}
\end{figure}

\begin{figure}[h]
    \centering
	\includegraphics[width=0.49\textwidth]{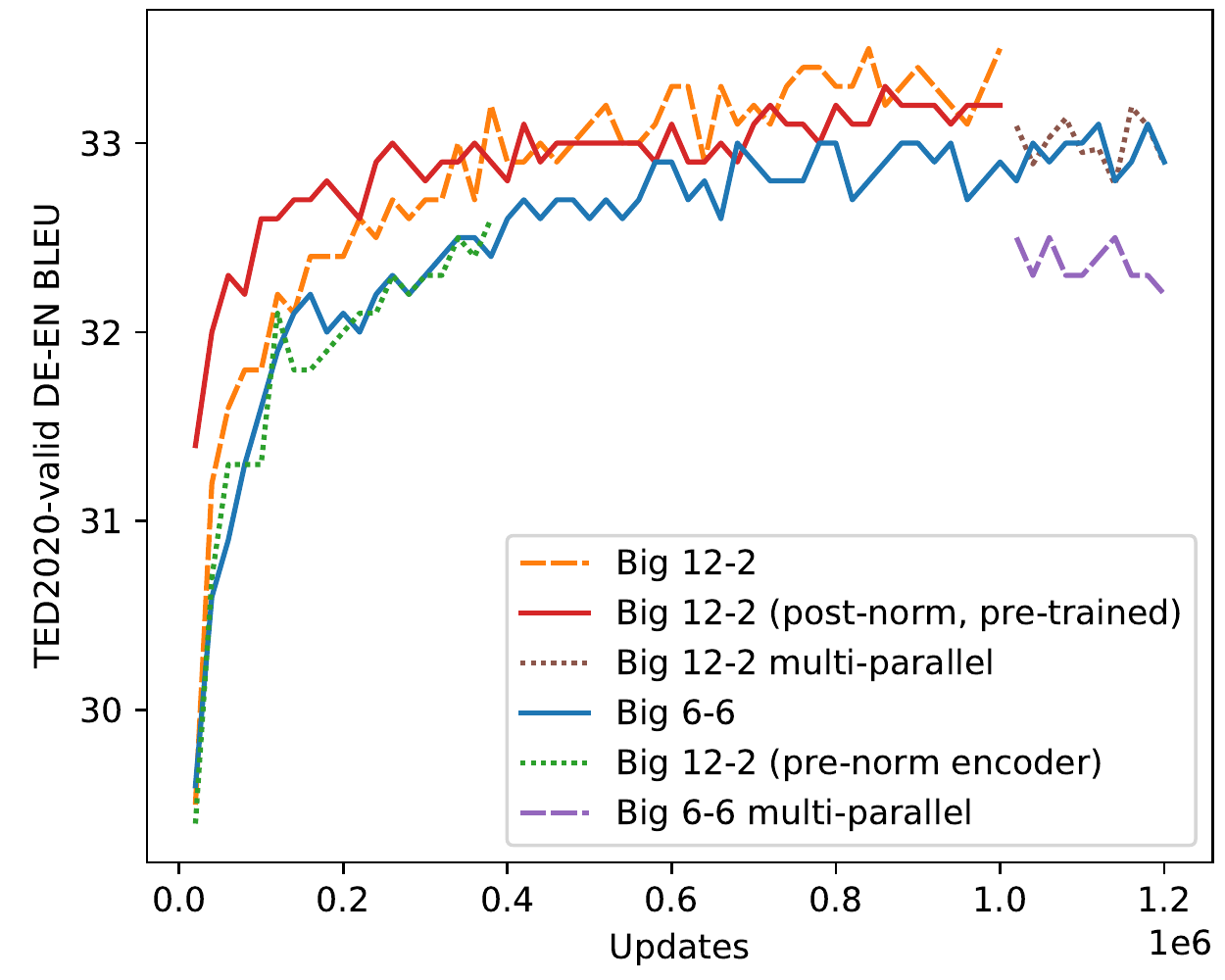}
	\includegraphics[width=0.49\textwidth]{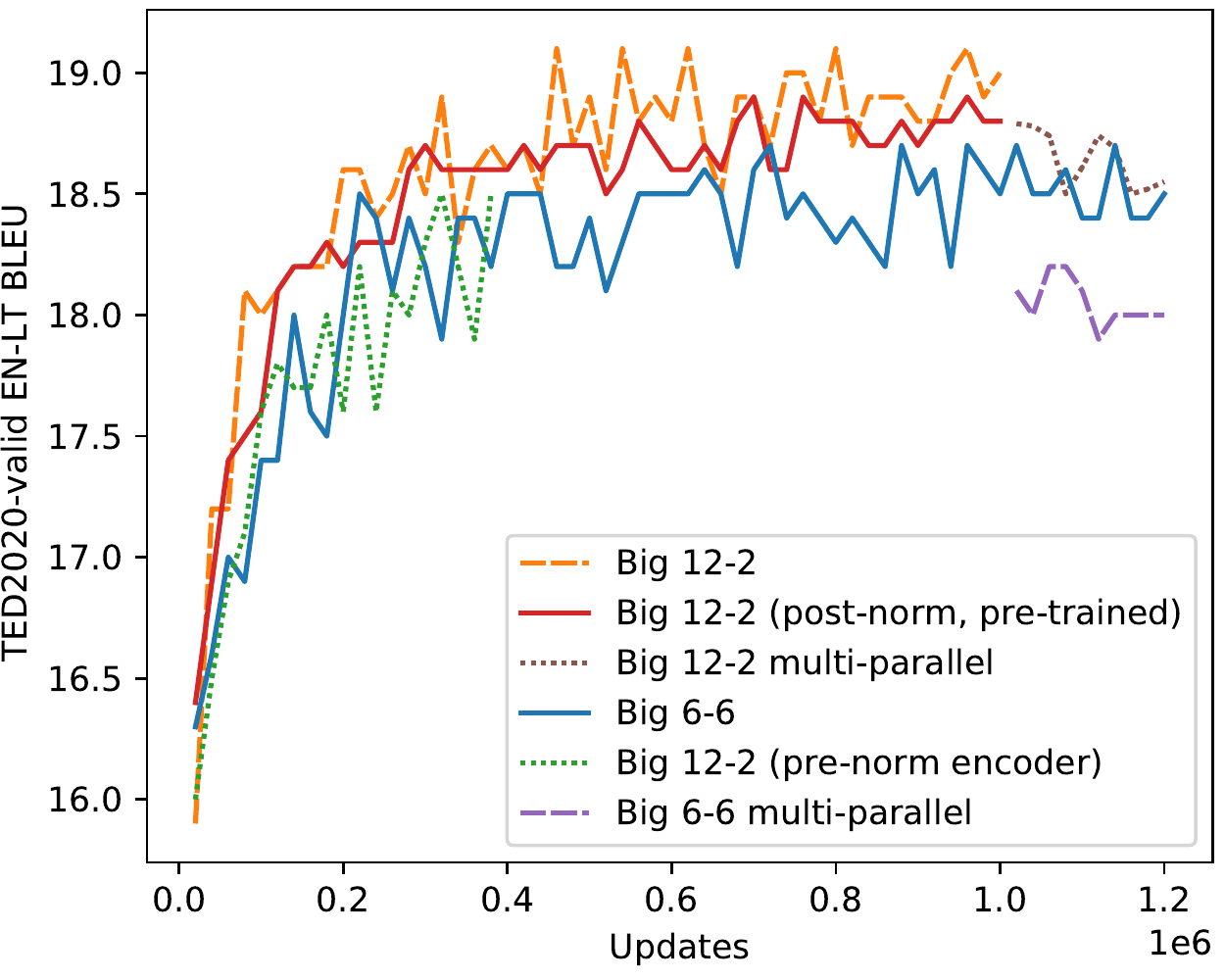}
	\includegraphics[width=0.49\textwidth]{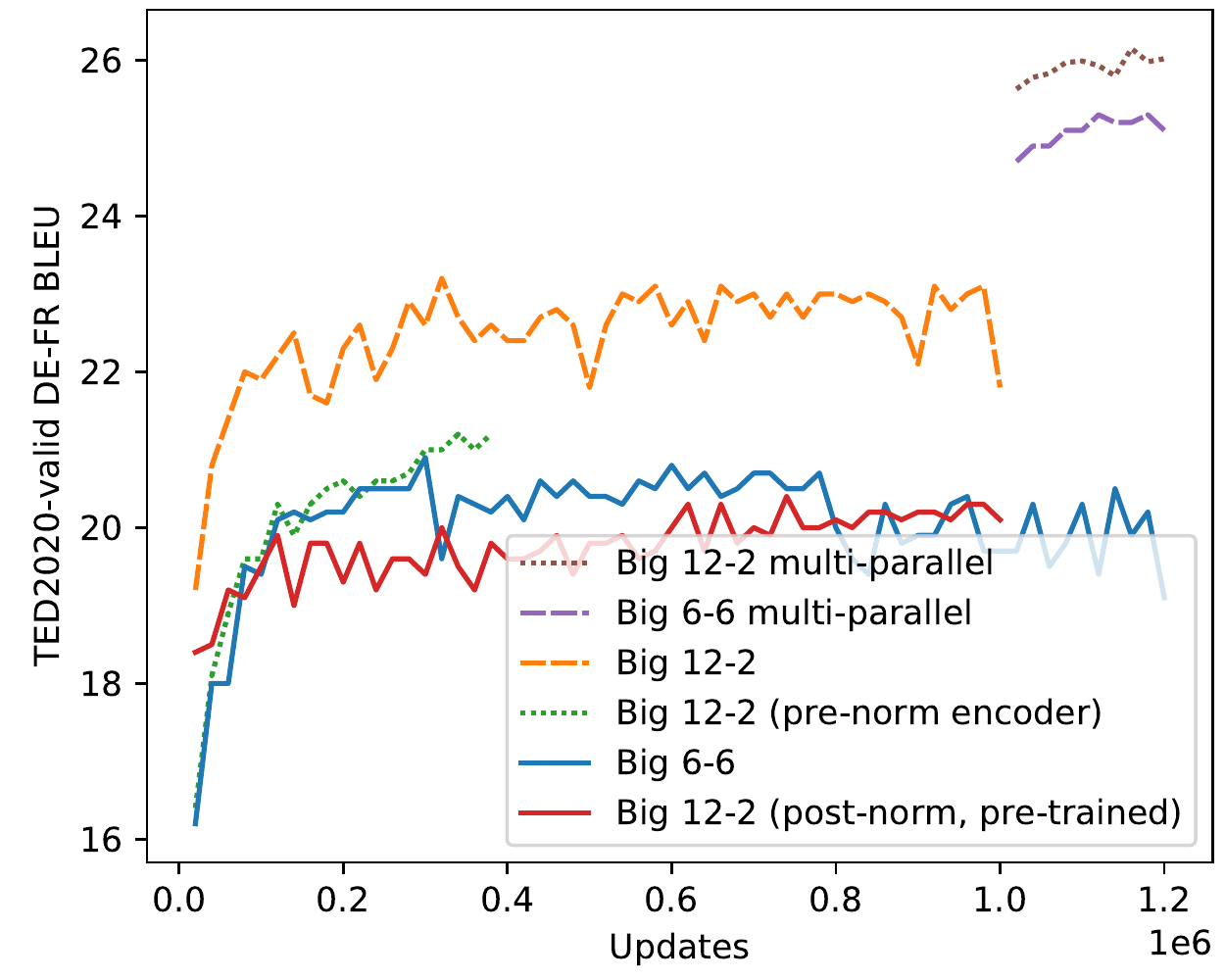}
	\includegraphics[width=0.49\textwidth]{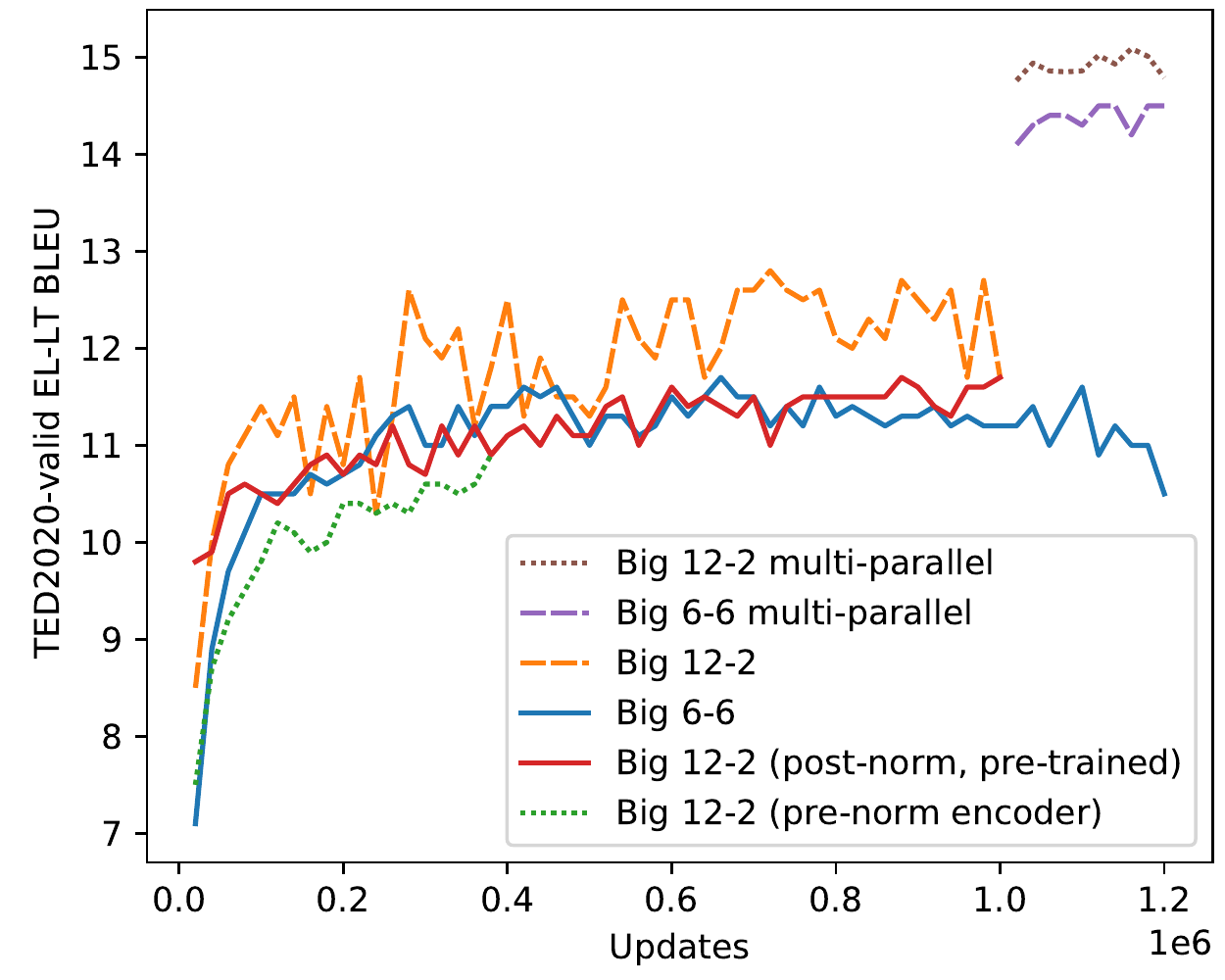}
	\caption{Training progress of English-centric and multi-parallel models trained on \textbf{ParaCrawl} (without BPE filtering). The names in the legend are sorted from highest to lowest BLEU score. The pre-trained 12-2 model was initialized with the 6-6 model's checkpoint at step 1M and fine-tuned for 1M more steps.}
\end{figure}

\begin{figure}[H]
    \centering
    \includegraphics[height=4.9cm]{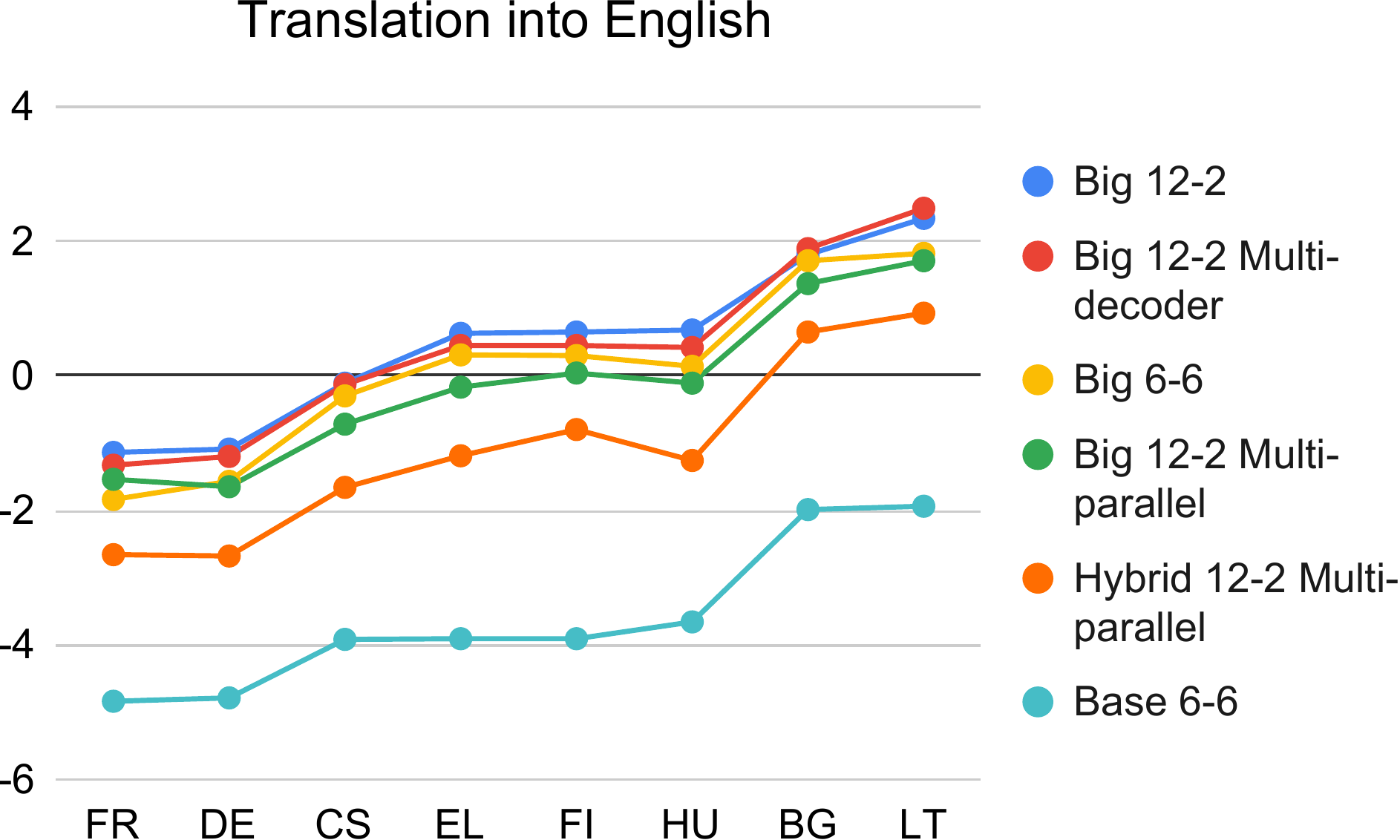}
    \includegraphics[height=4.9cm]{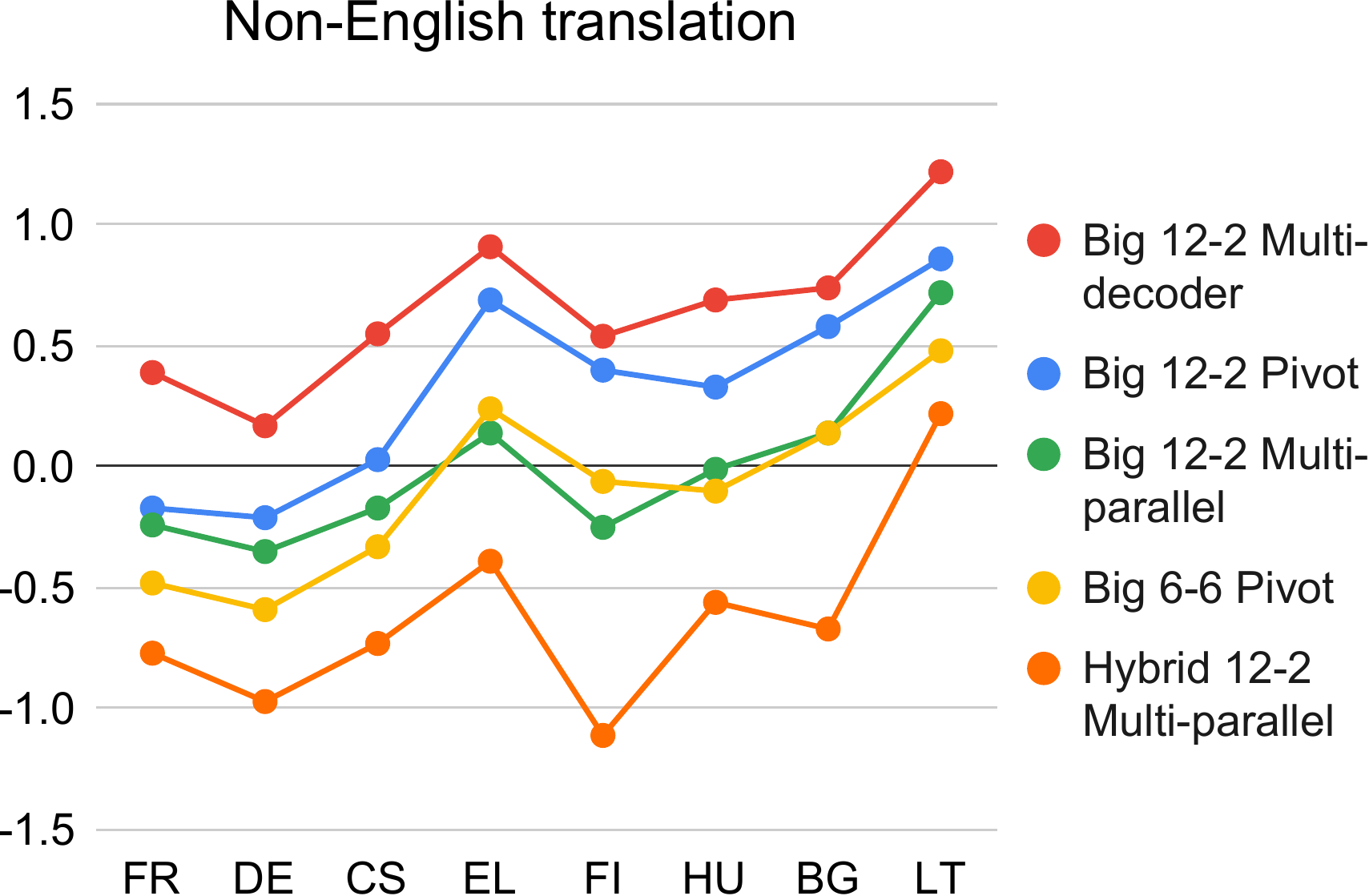}
    \caption{BLEU deltas on TED2020 test (in a subset of 8 languages) by our multilingual \textbf{ParaCrawl models} (\modelref{para_big_12_2}, \modelref{para_big_12_2_multidec}, \modelref{para_big_6_6}, \modelref{para_big_12_2_multipara_train_filter}, \modelref{para_hybrid_12_2}, \modelref{para_base_6_6}) compared to pivot translation through English with bilingual Big 6-6 baselines. The ``non-English'' score for language X is the average of test scores from all the other languages (except English) into X.
    The languages are sorted from highest-resource to lowest-resource (in terms of English-centric data amounts).}
    \label{fig:paracrawl_bleu_delta_non_english}
\end{figure}

\begin{table}[h]
	\centering
	\begin{tabular}{c|c|c|c|c|c|Hc}
	\multirow{2}{*}{Language} & \multirow{2}{*}{Code} & \multirow{2}{*}{Family} & \multicolumn{2}{c}{English-centric} & \multicolumn{3}{c}{Multi-parallel} \\
    & & & Lines & Trg. prob (T=5) & Lines & Trg. prob (T=5) & Trg. prob (T=2) \\
    \hline
    English & en & Germanic & 450,298,290 & 0.500 & 450,298,290 & 0.068 & 0.109 \\
    French & fr & Romance & 95,432,158 & 0.038 & 215,631,123 & 0.056 & 0.070 \\
    German & de & Romance & 76,490,492 & 0.036 & 192,673,679 & 0.056 & 0.068 \\
    Spanish & es & Romance & 72,973,508 & 0.036 & 191,713,598 & 0.056 & 0.068 \\
    Italian & it & Romance & 38,054,969 & 0.031 & 136,107,553 & 0.054 & 0.061 \\
    Portuguese & pt & Romance & 29,181,190 & 0.030 & 117,682,221 & 0.053 & 0.058 \\
    Dutch & nl & Germanic & 27,361,570 & 0.029 & 104,353,586 & 0.052 & 0.055 \\
    Norwegian & nb & Germanic & 15,384,700 & 0.026 & 65,367,888 & 0.049 & 0.045 \\
    Czech & cs & Slavic & 12,922,615 & 0.025 & 65,547,562 & 0.049 & 0.046 \\
    Polish & pl & Slavic & 12,877,872 & 0.025 & 69,265,211 & 0.050 & 0.047 \\
    Swedish & sv & Germanic & 10,969,372 & 0.025 & 60,160,949 & 0.048 & 0.044 \\
    Danish & da & Germanic & 9,792,687 & 0.024 & 61,276,295 & 0.049 & 0.044 \\
    Greek$^\star$ & el & Hellenic & 8,915,258 & 0.024 & 48,294,800 & 0.046 & 0.039 \\
    Finnish & fi & Uralic & 6,833,568 & 0.022 & 47,624,701 & 0.047 & 0.040 \\
    Croatian & hr & Slavic & 6,338,125 & 0.022 & 30,467,579 & 0.042 & 0.031 \\
    Hungarian & hu & Uralic & 6,294,289 & 0.022 & 42,527,237 & 0.046 & 0.037 \\
    Bulgarian$^\star$ & bg & Slavic & 6,098,653 & 0.022 & 36,835,445 & 0.044 & 0.035 \\
    Romanian & ro & Romance & 5,786,263 & 0.022 & 40,521,359 & 0.045 & 0.037 \\
    Slovak & sk & Slavic & 4,557,803 & 0.021 & 36,387,740 & 0.044 & 0.035 \\
    Lithuanian & lt & Baltic & 4,033,198 & 0.020 & 30,205,598 & 0.043 & 0.032 \\
    \hline
    Total & all & -- & 900,596,580 & 1.0 & 2,042,942,414 & 1.0 & 1.0 \\
    \hline
    Russian$^\star$ & ru & Slavic & 5,120,207 & -- & -- & -- & -- \\
    Latvian & lv & Baltic & 3,607,272 & -- & -- & -- & -- \\
	\end{tabular}
	\caption{Size of the \textbf{Top 20 \href{https://paracrawl.eu/v7-1}{ParaCrawl}} corpus and target language sampling probabilities in its English-centric setting and its multi-parallel setting. T = sampling temperature. English has 271,851,754 unique lines. $\star$: all languages use the latin script, except for Greek (Greek alphabet) and Bulgarian/Russian (Cyrillic). The average English sentence length is 19.0 words and 30.4 wordpieces.} 
	\label{tab:paracrawl_training_data}
\end{table}

\begin{table}[h]
    \centering
    \begin{tabular}{c@{\hspace{.2cm}}l|c c|c c}
        & \multicolumn{1}{c|}{Model} & \thead{BLEU \\ unk} & \thead{BLEU Consistency \\ unk }
        & \thead{BLEU \\ char} & \thead{BLEU Consistency \\ char }  \\
        \hline
        \modelref{para_big_6_6_multipara} & Big 6-6 & 24.2  & 73.3 & 19.5  & 54.2 \\
        \modelref{para_big_12_2_multipara} & Big 12-2 & \textbf{26.0} & \textbf{76.4} & \textbf{21.1} & \textbf{56.1}\\
        \modelref{para_big_12_2_multipara_test_filter} & Big 12-2  ($N_{test}=16$k) & 26.0 & 75.9 & 21.0  &56.0 \\
        \modelref{para_big_6_6_multipara_train_filter} & Big 6-6 ($N_{train}=8$k) & 24.4 & 74.4 & 19.6 & 54.5  \\
        \modelref{para_big_12_2_multipara_train_filter} & Big 12-2 ($N_{train}=8$k) & 25.6  & 73.7   & 20.9  &55.5 \\
        \modelref{para_hybrid_12_2} & Hybrid 12-2 ($N_{train}=8$k) & 24.8 & 75.0 & 20.4  & 55.3  \\
        \modelref{para_hybrid_12_3} & Hybrid 12-3 ($N_{train}=8$k) &  25.1 & 76.5 & 20.2  & 55.0   \\
    \end{tabular}
    \caption{Robustness evaluation on $\leftarrow$EN test sets of the multi-parallel \textbf{ParaCrawl models}. ``Unk'' adds one unknown symbol at a random position in each test sentence and ``char'' does 3 random character-level operations per sentence. ``BLEU consistency'' by \citet{niu2020} is a BLEU score between the translations of the clean and noisy versions of the same test set by a given model.}
    \label{tab:appendix:robustness}
\end{table}

\begin{table}
	\centering
	\begin{tabular}{|cc|}
		\hline
		Parameter name & Parameter value \\
		\hline
		share\_all\_embeddings & True \\
		arch & transformer \\
		lr\_scheduler & inverse\_sqrt \\
		optimizer & adam \\
		adam\_betas & 0.9,0.999 \\
		fp16 & True \\
		clip\_norm & 0.0 \\
		lr & 0.0005$^\star$ \\
		warmup\_updates & 4000 \\
		warmup\_init\_lr & 1e-07 \\
		criterion & label\_smoothed\_cross\_entropy \\
		label\_smoothing & 0.1 \\
		dropout & 0.3$^\star$ \\
		max\_tokens & 4000 \\
		max\_epoch & 120$^\star$ \\
		save-interval & 1 \\
		validate-interval & 1 \\
		keep-last-epochs & 1 \\
		update\_freq & 4$^\dagger$ \\
		\hline
		lang\_temperature & 5 \\
		decoder\_dropout & 0.3$^\star$ \\
		\hline
	\end{tabular}
	\vspace{-.2cm}
	\caption{fairseq hyper-parameters of the Base 6-6 \textbf{TED Talks models}. $\dagger$: we normalize this value by the number of GPUs to have a constant batch size. For instance, models trained on 4 GPUs use \texttt{update\_freq=1}. $\star$: as shown in Table~\ref{tab:ted_training_details}, the fine-tuned models use different values for these parameters. The Small 6-6 models use the \texttt{transformer\_iwslt\_de\_en} architecture.}
	\label{tab:ted_hyperparameters}
\end{table}

\begin{table}
	\centering
	\begin{tabular}{|cc|}
		\hline
	    max\_source\_positions & 256 \\
		max\_target\_positions & 256 \\
		share\_all\_embeddings & True \\
		arch & transformer\_vaswani\_wmt\_en\_de\_big \\
		lr\_scheduler & inverse\_sqrt \\
		optimizer & adam \\
		adam\_betas & 0.9,0.98 \\
		fp16 & True \\
		clip\_norm & 1.0 \\
		lr & 0.0005$^\star$ \\
		warmup\_updates & 4000 \\
		warmup\_init\_lr & 1e-07 \\
		criterion & label\_smoothed\_cross\_entropy \\
		label\_smoothing & 0.1 \\
		dropout & 0.1 \\
		max\_tokens & 8000 \\
		max\_update & 1000000$^\star$ \\
		save\_interval\_updates & 20000 \\
		validate\_interval\_updates & 20000 \\
		update\_freq & 32$^\dagger$ \\
		\hline
		lang\_temperature & 5$^\ddagger$ \\
		\hline
	\end{tabular}
	\vspace{-.2cm}
	\caption{fairseq hyper-parameters of the Big 6-6 \textbf{ParaCrawl models}. $\dagger$: we normalize this value by the number of GPUs to have a constant batch size. For instance, models trained on 4 GPUs use \texttt{update\_freq=8}. $\ddagger$: we use a temperature of 2 in the multi-parallel (or multi-decoder) finetuning stage. $\star$: as shown in Table~\ref{tab:paracrawl_training_details}, the fine-tuned models use different values for these parameters.}
	\label{tab:paracrawl_hyperparameters}
\end{table}

\begin{table}
	\centering
	\begin{tabular}{l|c|c|c|c|c}
		\multicolumn{1}{c|}{Models} & Initialized with & LR reset & Dropout & Data & Epochs \\
		\hline
		Base 6-6 (\modelref{ted_base_6_6}) & -- & -- & 0.3 & TED-EN & 120 \\
		Base 6-2 (\labelledmodelcounter{ted_base_6_2}) & -- & -- & 0.3 & TED-EN & 120 \\
		Base 12-2 (\modelref{ted_base_12_2}) & (\modelref{ted_base_6_6}) @60 & yes (0.0005) & 0.3 & TED-EN & 60 \\
		Base 6-6 Multi-parallel (\modelref{ted_base_6_6_multipara}, \modelref{ted_base_6_6_multipara_train_filter}) & (\modelref{ted_base_6_6}) @120 & no & 0.1 & TED-ALL & 10 \\
		Base 12-2 Multi-parallel (\modelref{ted_base_12_2_multipara}, \modelref{ted_base_12_2_multipara_train_filter}) & (\modelref{ted_base_12_2}) @60 & no & 0.1 & TED-ALL & 10 \\

		Hybrid Multi-parallel (\modelref{ted_hybrid_12_2}, \modelref{ted_hybrid_12_3}) & (\modelref{ted_base_12_2}) @60 & yes (0.0003) & 0.1 & TED-ALL & 10 \\		
		Base 12-2 Multi-decoder (\modelref{ted_base_12_2_multidec}) & (\modelref{ted_base_12_2}) @60 & yes (0.0001) & 0.1 / 0.3$^\star$ & TED-ALL & 10 \\
	\end{tabular}
	\caption{Details about multi-stage training of \textbf{TED Talks} models. $\star$: different dropout values in the encoder and decoders.}
	\label{tab:ted_training_details}
\end{table}

\begin{table}
	\centering
	\begin{tabular}{l|c|c|c|c}
		\multicolumn{1}{c}{Models} & Initialized with & LR reset & Data & Updates \\
		\hline
		Big 6-6 (\modelref{para_big_6_6}) & -- & -- & Para-EN & 1M \\
		Big 6-2 (\labelledmodelcounter{para_big_6_2}) & -- & -- & Para-EN & 1M \\
		Big 12-2 (\modelref{para_big_12_2}) & -- & -- & Para-EN & 1M \\
		Big 6-6 Multi-parallel (\modelref{para_big_6_6_multipara}, \modelref{para_big_6_6_multipara_train_filter}) & (\modelref{para_big_6_6}) @1M & no & Para-ALL & 200k \\
		Big 12-2 Multi-parallel (\modelref{para_big_12_2_multipara}, \modelref{para_big_12_2_multipara_train_filter}) & (\modelref{para_big_12_2}) @1M & no & Para-ALL & 200k \\
		Hybrid Multi-parallel (\modelref{para_hybrid_12_2}, \modelref{para_hybrid_12_3}) & (\modelref{para_big_12_2}) @1M & yes (0.0005) & Para-ALL & 200k \\
		Big 12-2 Multi-decoder (\modelref{para_big_12_2_multidec}) & (\modelref{para_big_12_2}) @1M & yes (0.0005) & Para-ALL & 200k \\
	\end{tabular}
	\caption{Details about multi-stage training of \textbf{ParaCrawl models}.}
	\label{tab:paracrawl_training_details}
\end{table}

\begin{table}
	\centering
	\begin{tabular}{l|c|c}
		\multicolumn{1}{c}{Models} & Params without embeddings (M) & Embeddings (M) \\
		\hline
		Base 6-6 (\modelref{ted_base_6_6}, \modelref{ted_base_6_6_multipara}) & 44.1 (18.9 + 25.2) & 36.0 \\
		Base 6-2 (\modelref{ted_base_6_2} & 27.3 (18.9 + 8.4) & 36.0 \\
		Base 12-2 (\modelref{ted_base_12_2}, \modelref{ted_base_12_2_multipara}) & 46.2 (37.8 + 8.4) & 36.0 \\
		Base 12-2 ($K=10$) (\modelref{ted_base_12_2_multipara_test_filter}) & 46.2 & 36.0 + 76.2 \\
		Base 12-2 ($N=4$k) (\modelref{ted_base_12_2_multipara_test_filter_more}, \modelref{ted_base_12_2_multipara_train_filter}) & 46.2 & 36.0 + 45.4 \\
		Hybrid 12-2 ($N=4$k) (\modelref{ted_hybrid_12_2}) & 43.6 (37.8 + 5.8) & 36.0 + 45.4 \\
		Hybrid 12-3 ($N=4$k) (\modelref{ted_hybrid_12_3}) & 46.8 (37.8 + 8.9) & 36.0 + 45.4 \\
		Multi-decoder Base 12-2 ($N=4$k) (\modelref{ted_base_12_2_multidec}) & 206.0 (37.8 + 20$\times$8.4) & 36.0 + 45.4 \\
	\end{tabular}
	\caption{Number of parameters in the \textbf{TED Talks models}.}
	\label{tab:ted_model_sizes}
\end{table}

\begin{table}
	\centering
	\begin{tabular}{l|c|c}
		\multicolumn{1}{c}{Models} & Params without embeddings (M) & Embeddings (M) \\
		\hline
		Big 6-6 (\modelref{para_big_6_6}, \modelref{para_big_6_6_multipara}) & 176.4 (75.6 + 100.8) & 70.7 \\
		Big 6-2 (\modelref{para_big_6_2}) & 109.2 (75.6 + 33.6) & 70.7 \\
		Big 12-2 (\modelref{para_big_12_2}, \modelref{para_big_12_2_multipara}) & 184.7 (151.1 + 33.6) & 70.7 \\
		Big 12-2 ($N=16$k) (\modelref{para_big_12_2_multipara_test_filter}) & 184.7 & 70.7 + 332.3 \\
		Big 12-2 ($N=8$k) (\modelref{para_big_12_2_multipara_test_filter_more}, \modelref{para_big_12_2_multipara_train_filter}) & 184.7 & 70.7 + 172.2 \\
		Hybrid 12-2 ($N=8$k) (\modelref{para_hybrid_12_2}) & 174.2 (151.1 + 23.1) & 70.7 + 172.2 \\
		Hybrid 12-3 ($N=8$k) (\modelref{para_hybrid_12_3}) & 186.8 (151.1 + 35.7) & 70.7 + 172.2 \\
		Multi-decoder Big 12-2 ($N=8$k) (\modelref{para_big_12_2_multidec}) & 823.0 (151.2 + 20$\times$33.6) & 70.7 + 172.2 \\
	\end{tabular}
	\caption{Number of parameters in the \textbf{ParaCrawl models}.}
	\label{tab:paracrawl_model_sizes}
\end{table}


\begin{table}
	\centering
	\begin{tabular}{c|c|c|c|c|c|c}
		\multirow{3}{*}{Time (s)} & \multirow{3}{*}{Parameters} & \multicolumn{3}{c|}{English-centric} & \multicolumn{2}{c}{Multi-parallel ($N_{train}=4$k)} \\
		& & Base 6-6 & Base 6-2 & Base 12-2 & Base 12-2 & Hybrid 12-2 \\
		& & (\modelref{ted_base_6_6}) & -- & (\modelref{ted_base_12_2}) & (\modelref{ted_base_12_2_multipara_train_filter}) & (\modelref{ted_hybrid_12_2}) \\
		\hline
		\multirow{4}{*}{Total} & beam=1 bs=1 & 18259 & 8455 & 9177 & 8483 & 5227 \\
		& beam=5 bs=1 & 34216 & 18215 & 18861 & 15351 & 10617 \\
		& beam=1 bs=64 & 1364 & 555 & 551 & 622 & 273 \\
		\cline{2-2}
		& \multirow{7}{*}{beam=5 bs=64} & 2330 & 1334 & 1329 & 1050 & 691 \\
		\cline{1-1}
		\cline{3-7}
		Encoder & & 15 & 15 & 29 & 30 & 29 \\
		\cline{1-1}
		\cline{3-7}
		Decoder & & 1435 & 554 & 545 & 548 & 234 \\
		Self-attn / RNN & & 477 & 162 & 159 & 168 & 71 \\
		Cross-attn & & 399 & 133 & 132 & 139 & 60 \\
		Softmax & & 49 & 48 & 48 & 21 & 21 \\
		\cline{1-1}
		\cline{3-7}
		Beam top-k & & 350 & 341 & 334 & 38 & 38 \\
	\end{tabular}
	\caption{Time benchmark across different \textbf{TED Talks model} sizes and decoding settings. Time in seconds spent decoding the concatenation of all X$\to$EN TED valid sets (averages over 3 runs). Note that to mimic a true online setting, no sorting by length is applied (i.e., \texttt{buffer\_size=batch\_size}). We modify fairseq's code to avoid the slow beam search code when \texttt{beam=1} (which unnecessarily computes and stores log probabilities). We see that in this setting, vocabulary size has a minor impact on speed.}
	\label{tab:ted_speed_benchmark}
\end{table}

\begin{table}
	\centering
	\begin{tabular}{c|c|c|c|c|c|c}
		\multirow{3}{*}{Time (s)} & \multirow{3}{*}{Parameters} & \multicolumn{3}{c|}{English-centric} & \multicolumn{2}{c}{Multi-parallel ($N_{train}=8$k)} \\
		& & Big 6-6 & Big 6-2 & Big 12-2 & Big 12-2 & Hybrid 12-2 \\
		& & (\modelref{para_big_6_6}) & (\modelref{para_big_6_2}) & (\modelref{para_big_12_2}) & (\modelref{para_big_12_2_multipara_train_filter}) & (\modelref{para_hybrid_12_2}) \\
		\hline
		\multirow{4}{*}{Total} & beam=1 bs=1 & 14464 & 6131 & 6535 & 6544 & 3519 \\
		& beam=5 bs=1 & 23990 & 12646 & 13264 & 10873 & 7034 \\
		& beam=1 bs=64 & 912 & 375 & 402 & 397 & 195 \\
		\cline{2-2}
		& \multirow{7}{*}{beam=5 bs=64} & 1492 & 854 & 902 & 708 & 495 \\
		\cline{1-1}
		\cline{3-7}
		Encoder & & 21 & 21 & 41 & 41 & 41 \\
		\cline{1-1}
		\cline{3-7}
		Decoder & & 899 & 350 & 364 & 343 & 155 \\
		Self-attn / RNN & & 296 & 101 & 105 & 105 & 44 \\
		Cross-attn & & 257 & 85 & 89 & 88 & 42 \\
		Softmax & & 39 & 38 & 39 & 19 & 20 \\
		\cline{1-1}
		\cline{3-7}
		Beam top-k & & 207 & 204 & 209 & 40 & 40 \\
	\end{tabular}
	\caption{Time benchmark across different \textbf{ParaCrawl model} sizes and decoding settings. Time in seconds spent decoding the concatenation of all $\to$EN TED2020-valid sets (averages over 3 runs).}
	\label{tab:para_speed_benchmark}
\end{table}

\begin{table}
	\centering
	\begin{tabular}{c@{\hspace{.2cm}}c|c|c|c|c}
		& Model & $\to$EN & $\leftarrow$EN & $/$ EN & WPS \\
		\hline
		& \multicolumn{5}{c}{English-centric} \\
		\hline
		\modelref{ted_base_6_6} & Base 6-6 & 31.8 / 30.5 & 24.2 / 23.6 & 13.5 / 13.2  & 724 / 385 \\
		\modelref{ted_base_12_2} & Base 12-2 & 33.6 / 32.9 & 24.3 / 24.2 & 14.1 / 14.1 & 1321 / 765\\
		\hline
		& \multicolumn{5}{c}{+ Multi-parallel} \\
		\hline
		\modelref{ted_base_6_6_multipara} & Base 6-6 & 32.8 / 31.5  & 24.3 / 23.4  & 16.3 / 15.6  & 760 / 390\\
		\modelref{ted_base_12_2_multipara} & Base 12-2 & 33.5 / 32.7 & 24.5 / 24.2 & 16.3 / 16.0 & 1258 / 723\\
		\modelref{ted_hybrid_12_2} & Hybrid 12-2 & 32.8 / 31.7 & 23.5 / 23.6  & 16.1 / 15.6 & 2546 / 1403\\
		\modelref{ted_hybrid_12_3} & Hybrid 12-3 & 32.9 / 31.6  & 23.7 / 23.6 & 16.1 / 15.6 & 2279 / 1338 \\
	\end{tabular}
	\caption{\textbf{TED Talks} experiments on another MT framework show that our results are reproducible. The first number in each cell is the value obtained with fairseq and the second number is obtained with our internal TensorFlow implementation. Both implementations share the same hyper-parameters, with one notable exception: in the TensorFlow implementation, source/target embeddings are not shared. Additionally, the Hybrid models trained with fairseq use train-time BPE filtering, while the TensorFlow models do not.}
	\label{tab:reproducibility}
\end{table}  

\newpage

\begin{figure}
    \centering
    \includegraphics[height=4.5cm]{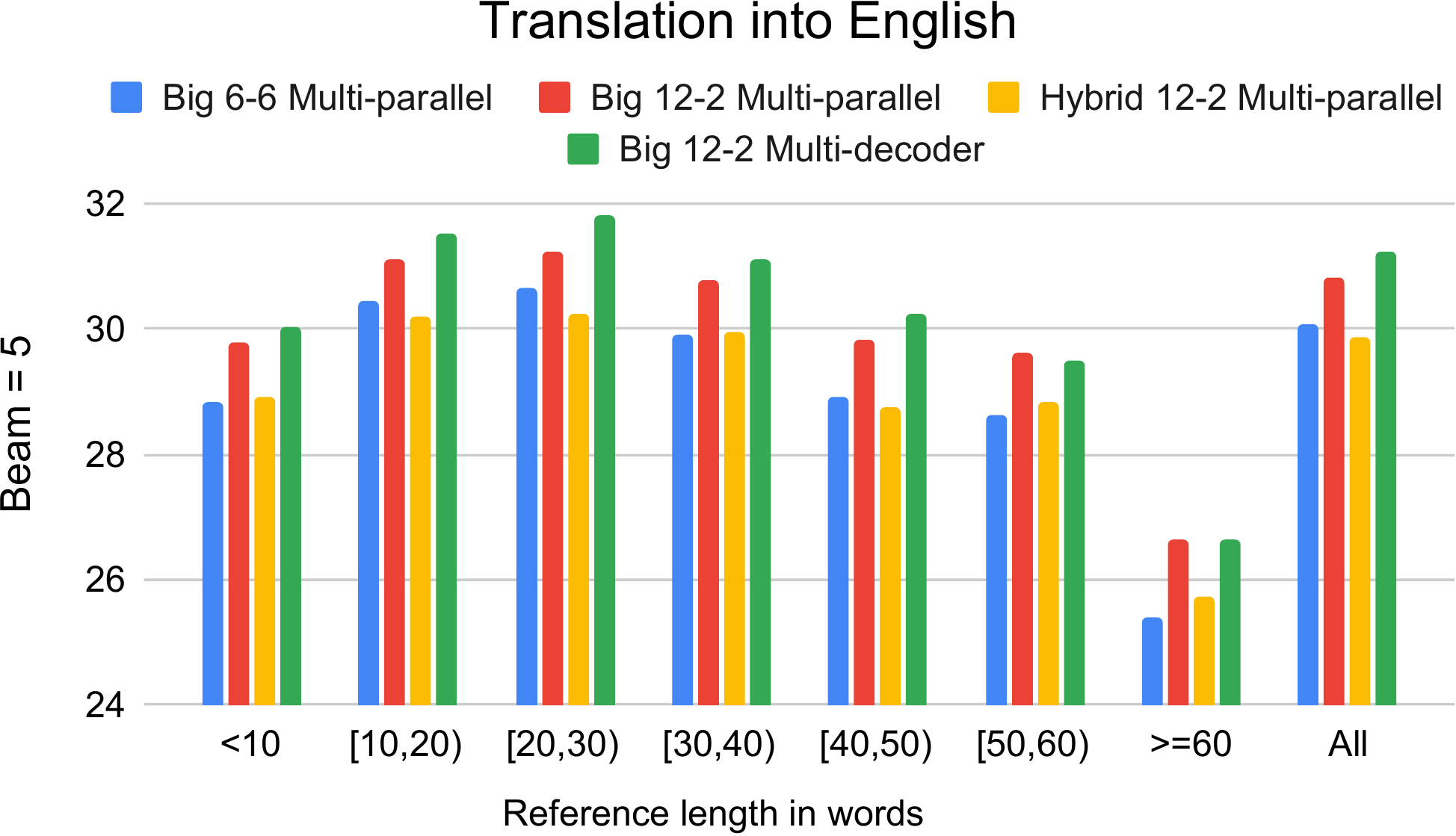}
    \includegraphics[height=4.5cm]{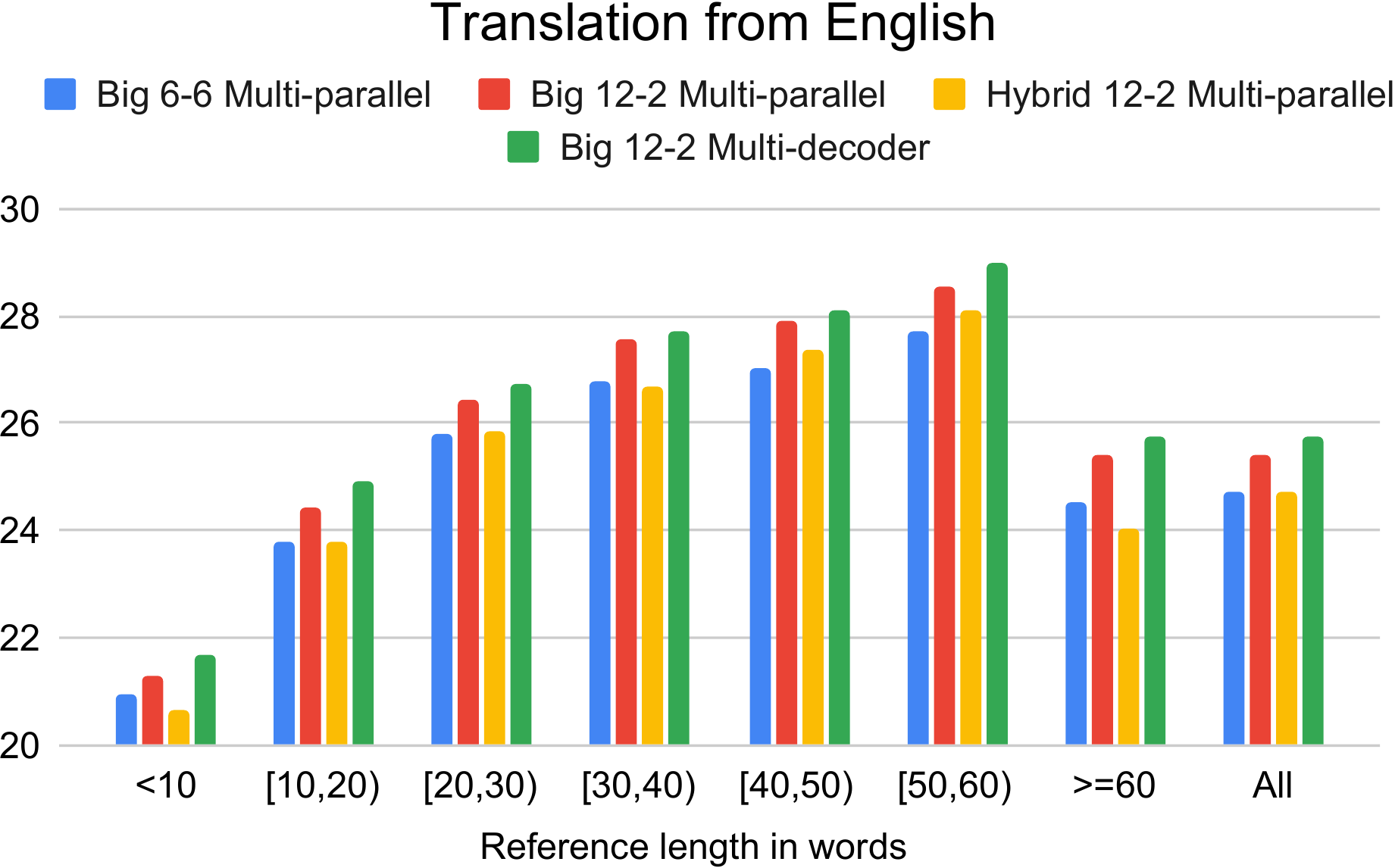} \\[.5cm]
    \includegraphics[height=3.6cm]{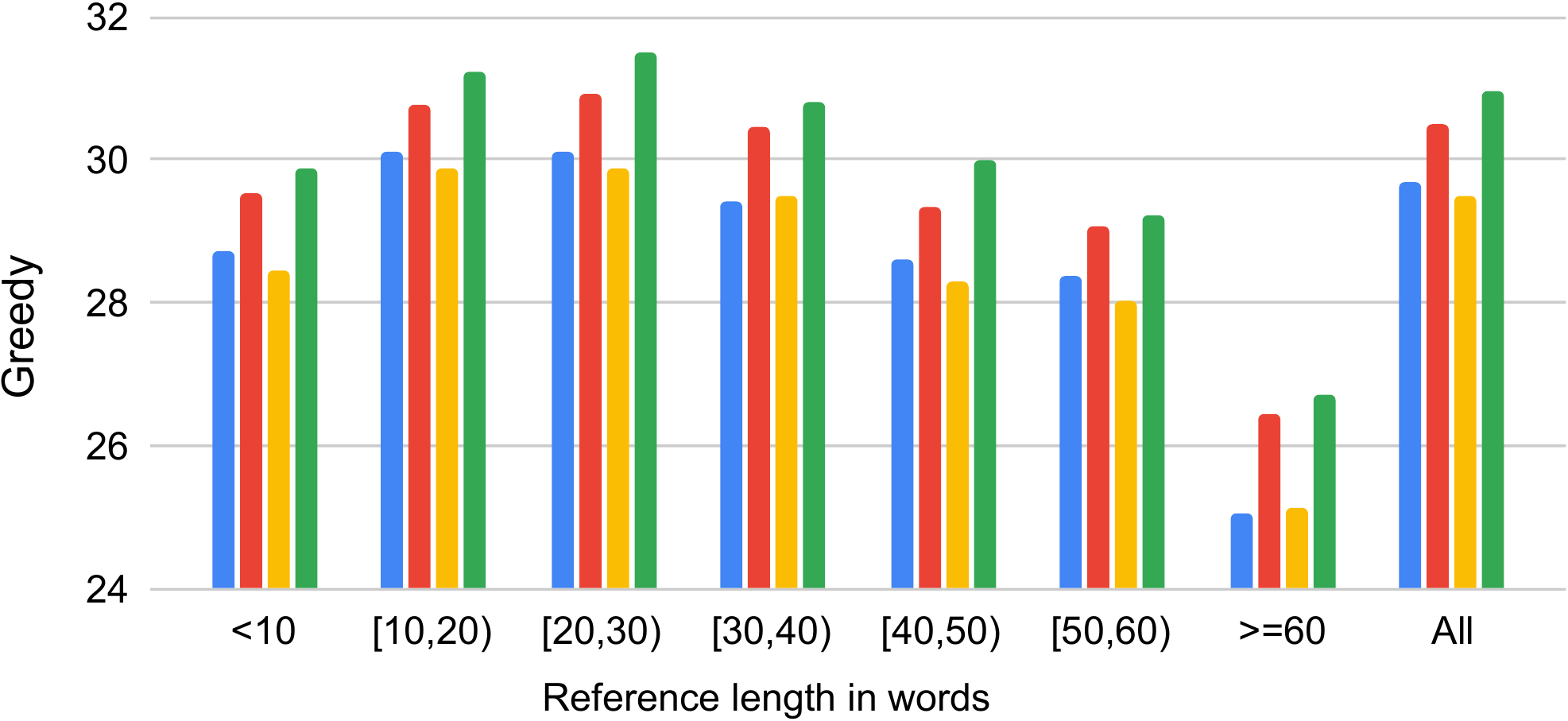}
    \includegraphics[height=3.6cm]{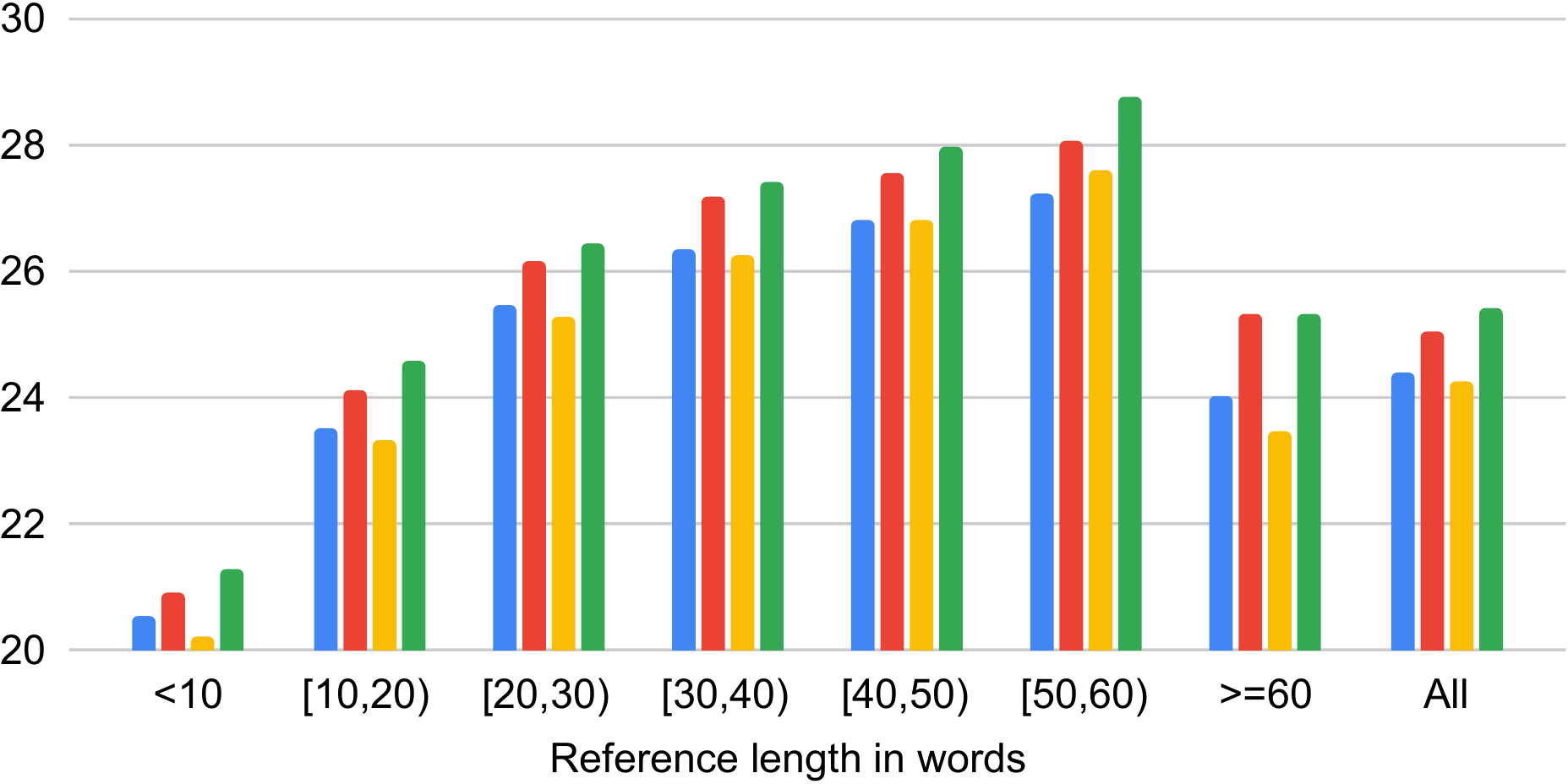}
    \caption{BLEU scores on $\leftrightarrow$EN TED2020 test by the \textbf{ParaCrawl models} with beam search or greedy search (\modelref{para_big_6_6_multipara_train_filter}, \modelref{para_big_12_2_multipara_train_filter}, \modelref{para_hybrid_12_2}, \modelref{para_big_12_2_multidec}) according to sentence length. Top: beam search. Bottom: greedy search. Left: $\to$EN translation. Right: $\leftarrow$EN translation.}
    \label{fig:bleu_by_length_more}
\end{figure}

\begin{table}[t]
    \centering
    \begin{tabular}{c@{\hspace{.2cm}}c|ccc|ccc|HHHHHHc}
        & \multirow{2}{*}{Model} & \multicolumn{3}{c|}{Test BLEU} & \multicolumn{3}{c|}{Test chrF} & \multicolumn{3}{H}{Valid BLEU} & \multicolumn{3}{H}{Valid chrF} & \multirow{2}{*}{WPS} \\

		& & $\to$EN & $\leftarrow$EN & $/$ EN & $\to$EN & $\leftarrow$EN & $/$ EN & $\to$EN & $\leftarrow$EN & $/$ EN & $\to$EN & $\leftarrow$EN & $/$ EN & \\
		\cline{2-15}
		& \multicolumn{14}{c}{SOTA \citep{philip-etal-2020-monolingual}} \\
		\cline{2-15}
		& Bilingual & 32.4 & 24.4 & 15.0 & -- & -- & -- & -- & -- & -- & -- & -- & -- & -- \\   
		& Best multi & 32.3 & 24.1 & 15.8 & -- & -- & -- & -- & -- & -- & -- & -- & -- & -- \\        
		\cline{2-15}
		& \multicolumn{14}{c}{English-centric} \\
		\cline{2-15}
		\modelref{ted_small_6_6} & Small 6-6 & 31.6 & 23.1 & 11.6 (14.4) & .527 & .474 & .342 (.373) & 31.7 & 23.1 & 11.6 (14.5) & .526 & .473 & .341 (.373) & 703 \\ 
		\modelref{ted_base_6_6} & Base 6-6 & 31.8 & 24.2 & 13.5 (15.0) & .529 & .486 & .370 (.381) & 32.2 & 24.3 & 13.5 (15.1) & .530 & .487 & .371 (.382) & 753 \\ 
		\modelref{ted_base_6_6_lang_as_bos} & (\modelref{ted_base_6_6}) + dec. lang code & 32.0 & 23.9 & 5.4 (14.9) & .532 & .484 & .186 (.381) & 32.3 & 23.9 & 5.5 (15.1) & .532 & .483 & .186 (.382) & 766 \\ 
		\modelref{ted_base_6_2} & Base 6-2 & 32.5 & 23.3 & 12.7 (14.6) & .536 & .475 & .358 (.377) & 32.6 & 23.4 & 12.7 (14.7) & .535 & .475 & .358 (.377) & 1271 \\ 
		\modelref{ted_base_12_2} & Base 12-2, init with (\modelref{ted_base_6_6}) & 33.6 & 24.3 & 14.1 (15.4) & .548 & .487 & .378 (.389) & 33.7 & 24.4 & 14.1 (15.5) & .548 & .486 & .378 (.389) & 1287 \\ 
		\modelref{ted_base_12_2_prenorm} & Base 12-2, pre-norm & 33.1 & 24.2 & 13.2 (15.3) & .544 & .487 & .368 (.388) & 33.3 & 24.1 & 13.1 (15.3) & .544 & .486 & .368 (.388) & 1312 \\ 
		\modelref{ted_base_12_2_enc_prenorm} & Base 12-2, enc. pre-norm & 33.3 & 24.3 & 12.7 (15.3) & .546 & .487 & .359 (.389) & 33.5 & 24.3 & 12.6 (15.4) & .546 & .487 & .357 (.389) & 1240 \\ 
		\modelref{ted_base_12_2_lang_as_bos} & (\modelref{ted_base_12_2}) + dec. lang code & 33.4 & 22.8 & 0.7 (14.7) & .547 & .472 & .102 (.380) & 33.6 & 22.8 & 1.0 (14.7) & .547 & .471 & .104 (.380) & 1307 \\ 
		\labelledmodelcounter{ted_base_6_6_top59} & (\modelref{ted_base_6_6}) + TED-59 & 25.5 & 19.9 & 9.0 (11.5) & .455 & .423 & .282 (.324) & 25.7 & 20.0 & 8.9 (11.6) & .456 & .422 & .279 (.324) & 715 \\ 
		\cline{2-15}
		& \multicolumn{14}{c}{+ Multi-parallel} \\
		\cline{2-15}
		\modelref{ted_base_6_6_multipara} & Base 6-6 & 32.8 & 24.3 & 16.3 & .540 & .482 & .395 & 33.1 & 24.3 & 16.3 & .541 & .482 & .395 & 732 \\ 
		\labelledmodelcounter{ted_base_6_6_multipara_test_filter} & (\modelref{ted_base_6_6_multipara}) + $K_{test}=10$ & 32.7 & 24.1 & 16.3 & .539 & .481 & .394 & 33.0 & 24.1 & 16.2 & .539 & .480 & .394 & 826 \\ 
		\labelledmodelcounter{ted_base_6_6_multipara_test_filter_more} & (\modelref{ted_base_6_6_multipara}) + $N_{test}=4$k & 30.8 & 22.3 & 15.2 & .519 & .463 & .382 & 31.1 & 22.3 & 15.2 & .519 & .463 & .382 & 868 \\ 
		\modelref{ted_base_12_2_multipara} & Base 12-2 & 33.5 & 24.5 & 16.3 & .546 & .485 & .395 & 33.8 & 24.4 & 16.4 & .547 & .484 & .395 & 1203 \\ 
		\modelref{ted_base_12_2_multipara_test_filter} & (\modelref{ted_base_12_2_multipara}) + $K_{test}=10$ & 33.4 & 24.3 & 16.2 & .545 & .484 & .394 & 33.7 & 24.2 & 16.3 & .546 & .483 & .394 & 1539 \\ 
		\modelref{ted_base_12_2_multipara_test_filter_more} & (\modelref{ted_base_12_2_multipara}) + $N_{test}=4$k & 31.5 & 22.5 & 15.2 & .523 & .466 & .382 & 31.8 & 22.3 & 15.2 & .525 & .465 & .382 & 1457 \\ 
		\labelledmodelcounter{ted_hybrid_12_2_multipara} & Hybrid 12-2 & 32.9 & 23.7 & 16.2 & .544 & .476 & .394 & 33.2 & 23.7 & 16.2 & .545 & .475 & .394 & 1724 \\ 
		\labelledmodelcounter{ted_hybrid_12_3_multipara} & Hybrid 12-3 & 32.9 & 23.8 & 16.2 & .543 & .476 & .395 & 33.3 & 23.8 & 16.3 & .544 & .476 & .395 & 1533 \\ 
		\cline{2-15}
		& \multicolumn{14}{c}{+ BPE filtering ($N_{train}=4$k)} \\
		\cline{2-15}
		\modelref{ted_base_6_6_multipara_train_filter} & Base 6-6 & 32.9 & 24.2 & 16.3 & .541 & .481 & .394 & 33.2 & 24.2 & 16.2 & .541 & .481 & .394 & 789 \\ 
        \labelledmodelcounter{ted_base_6_6_lang_as_bos_multipara_train_filter} & (\modelref{ted_base_6_6_multipara_train_filter}) + dec. lang code & 32.7 & 23.9 & 16.2 & .540 & .478 & .393 & 33.0 & 24.0 & 16.2 & .540 & .479 & .394 & 891 \\ 
        \labelledmodelcounter{ted_base_6_2_multipara_train_filter} & Base 6-2 & 32.5 & 23.4 & 15.6 & .537 & .472 & .386 & 32.7 & 23.4 & 15.6 & .538 & .472 & .386 & 1595 \\ 
		\modelref{ted_base_12_2_multipara_train_filter} & Base 12-2 & 33.3 & 24.3 & 16.3 & .546 & .481 & .395 & 33.4 & 24.4 & 16.4 & .546 & .481 & .395 & 1552 \\ 
		\labelledmodelcounter{ted_base_12_2_lang_as_bos_multipara_train_filter} & (\modelref{ted_base_12_2_multipara_train_filter}) + dec. lang code &32.5 & 23.0 & 15.7 & .540 & .471 & .389 & 32.7 & 23.0 & 15.7 & .540 & .470 & .389 & 1514 \\ 
		\modelref{ted_hybrid_12_2} & Hybrid 12-2 & 32.8 & 23.5 & 16.1 & .543 & .474 & .393 & 33.1 & 23.6 & 16.1 & .543 & .474 & .393 & \textbf{2422} \\ 
		\modelref{ted_hybrid_12_3} & Hybrid 12-3 & 32.9 & 23.7 & 16.1 & .543 & .475 & .394 & 33.2 & 23.7 & 16.2 & .544 & .474 & .393 & 2145 \\ 
		\cline{2-15}
        & \multicolumn{14}{c}{+ Multi-decoder} \\
        \cline{2-15}
        \modelref{ted_base_6_2_multidec} & (\modelref{ted_base_6_6}) $\to$ 6-2 & 33.0 & 24.2 & 16.0 & .543 & .480 & .391 & 33.3 & 24.2 & 16.0 & .543 & .479 & .391 & 1608 \\ 
        \labelledmodelcounter{ted_base_6_2_multidec_from_6_2} & (\modelref{ted_base_6_2}) $\to$ 6-2 & 33.3 & 24.5 & 16.2 & .545 & .482 & .394 & 33.6 & 24.4 & 16.3 & .546 & .482 & .394 & 1548 \\ 
		\modelref{ted_base_12_2_multidec} & (\modelref{ted_base_12_2}) $\to$ 12-2 & \textbf{33.8} & \textbf{25.1} & \textbf{16.7} & .551 & .490 & .401 & 34.0 & 25.1 & 16.8 & .552 & .489 & .402 & 1614 \\ 
 		\labelledmodelcounter{ted_base_12_2_multidec_no_lang_code} & (\modelref{ted_base_12_2_multidec}) w/o lang code & 33.6 & 24.3 & 16.6 & .550 & .481 & .399 & 33.9 & 24.4 & 16.6 & .550 & .481 & .399 & 1567 \\ 
	\end{tabular}
	\caption{Test BLEU, chrF scores and decoding speed of \textbf{TED Talks models} of various depths. SOTA's ``best multi'' is a Transformer Small 6-6 multi-parallel model with adapter layers. WPS: speed in words per second for $\to$EN translation. Scores in parentheses are obtained by pivot translation through English. (\modelref{ted_base_12_2_multidec_no_lang_code}) does not use any language code during the multi-decoder finetuning stage.}
	\label{tab:ted_scores_more}
\end{table}

\begin{table}[t]
 	\centering
	\begin{tabular}{c@{\hspace{.2cm}}c|HHHccc|HHHccc|c}
		& \multirow{2}{*}{Model} & \multicolumn{3}{H}{TED2020 test BLEU} & \multicolumn{3}{c|}{TED2020 test chrF} & \multicolumn{3}{H}{TED2020 valid chrF} & \multicolumn{3}{c|}{FLORES devtest spBLEU} & \multirow{2}{*}{WPS} \\
		& & $\to$EN & $\leftarrow$EN & $/$ EN & $\to$EN & $\leftarrow$EN & $/$ EN & $\to$EN & $\leftarrow$EN & $/$ EN & $\to$EN & $\leftarrow$EN & $/$ EN & \\
		\cline{2-15}
		& \citet{goyal2021flores101} & -- & -- & -- & -- & -- & -- & -- & -- & -- & 32.4 & 31.9 & 25.7 & -- \\ %
		\cline{2-15}
		& \multicolumn{14}{c}{English-centric} \\
		\cline{2-15}
		\modelref{para_base_6_6} & Base 6-6 & 31.4 & 26.0 & 13.3 (18.2) & .551 & .545 & .399 (.456) & .553 & .547 & .399 (.456) & 34.0 & 31.9 & 16.0 (22.5) & 656 \\ 
		\modelref{para_big_6_6} & Big 6-6 & 35.0 & 28.9 & 14.4 (21.1) & .580 & .569 & .401 (.483)
 & .582 & .571 & .400 (.483) & 38.8 & 36.4 & 18.5 (27.0) & 623 \\ 
		\modelref{para_big_6_2} & Big 6-2 & 33.9 & 28.4 & 12.2 (20.3) & .572 & .565 & .373 (.477)
& .575 & .567 & .372 (.477) & 37.7 & 35.3 & 15.3 (25.8) & 1077 \\ 
		\modelref{para_big_12_2} & Big 12-2 & 35.5 & 29.6 & 16.7 (21.5) & .585 & .575 & .435 (.488)
& .587 & .577 & .435 (.488) & 39.6 & 37.1 & 21.1 (27.6) & 1033 \\ 
        \labelledmodelcounter{para_wide_12_2} & Wide 12-2 & \textbf{36.1} & \textbf{30.4} & 13.9 (\textbf{22.2}) & \textbf{.590} & \textbf{.581} & .381 (\textbf{.494})
& \textbf{.592} & \textbf{.583} & .381 (\textbf{.494}) & \textbf{40.7} & \textbf{38.9} & 19.0 (\textbf{29.0}) & 870 \\ 
		\cline{2-15}
		& \multicolumn{14}{c}{+ Multi-parallel} \\
		\cline{2-15}
		\modelref{para_big_6_6_multipara} & Big 6-6 & 34.2 & 28.3 & 20.7 & .574 & .564 & .481
& .576 & .566 & .481 & 37.9 & 35.6 & 26.8 & 595 \\ 
		\labelledmodelcounter{para_big_6_6_multipara_T5} & Big 6-6 ($T=5$) & 34.0 & 28.3 & 20.7 & .573 & .564 & .481
& .575 & .565 & .481 & 37.6 & 35.6 & 26.8 & 608 \\ 
		\modelref{para_big_12_2_multipara} & Big 12-2 & 34.9 & 29.1 & 21.3 & .581 & .570 & .486
& .582 & .572 & .486 & 39.0 & 36.2 & 27.6 & 1030 \\ 
		\modelref{para_big_12_2_multipara_test_filter} & (\modelref{para_big_12_2_multipara}) + $N_{test}=16$k & 34.8 & 29.0 & 21.2 & .580 & .570 & .486
& .582 & .572 & .486 & 38.4 & 35.8 & 27.2 & 1328 \\ 
		\modelref{para_big_12_2_multipara_test_filter_more} & (\modelref{para_big_12_2_multipara}) + $N_{test}=8$k & 32.9 & 27.8 & 20.4 & .563 & .560 & .478
& .565 & .562 & .478 & 33.5 & 33.4 & 25.3 & 1283 \\ 
		\labelledmodelcounter{para_hybrid_12_3_no_filter} & Hybrid 12-3 & 34.2 & 28.4 & 20.9 & .576 & .564 & .482
& .578 & .566 & .482 & 38.2 & 35.4 & 26.9 & 1313 \\ 
		\cline{2-15}
		& \multicolumn{14}{c}{+ BPE filtering ($N_{train}=8$k)} \\
		\cline{2-15}
		\modelref{para_big_6_6_multipara_train_filter} & Big 6-6 & 34.0 & 28.4 & 20.7 & .572 & .564 & .480
& .575 & .566 & .480 & 37.5 & 35.5 & 26.6 & 679 \\ 
		\labelledmodelcounter{para_big_6_2_multipara_train_filter} & Big 6-2 & 32.9 & 27.8 & 19.8 & .565 & .558 & .472
& .567 & .560 & .472 & 36.2 & 34.0 & 25.1 & 1305 \\ 
		\modelref{para_big_12_2_multipara_train_filter} & Big 12-2 & 34.8 & 29.1 & 21.2 & .580 & .570 & .486
& .582 & .572 & .486 & 38.8 & 36.2 & 27.4 & 1261 \\ 
		\modelref{para_hybrid_12_2} & Hybrid 12-2 & 33.9 & 28.3 & 20.7 & .573 & .562 & .481
& .575 & .564 & .481 & 37.9 & 34.9 & 26.5 & 1796 \\ 
		\labelledmodelcounter{para_hybrid_12_2_longer} & (\modelref{para_hybrid_12_2}) + 200k steps & 34.2 & 28.4 & 20.9 & .576 & .564 & .483
& .578 & .566 & .483 & 38.2 & 34.9 & 26.8 & \textbf{1861} \\ 
 		\modelref{para_hybrid_12_3} & Hybrid 12-3 & 34.1 & 28.3 & 20.8 & .574 & .563 & .482
& .576 & .564 & .482 & 38.0 & 34.8 & 26.6 & 1659 \\ 
 		\labelledmodelcounter{para_hybrid_12_3_longer} & (\modelref{para_hybrid_12_3}) + 200k steps & 34.3 & 28.6 & 21.0 & .577 & .565 & .483
& .578 & .567 & .483 & 38.3 & 35.6 & 27.1 & 1770 \\ 
		\cline{2-15}
		& \multicolumn{14}{c}{+ Multi-decoder} \\
		\cline{2-15}
		\modelref{para_big_6_2_multidec} & (\modelref{para_big_6_6}) $\to$ 6-2 & 34.0 & 28.8 & 21.0 & .573 & .566 & .483
& .575 & .568 & .483 & 38.0 & 36.3 & 27.3 & 1333 \\ 
		\labelledmodelcounter{para_big_6_2_multidec_longer} & (\modelref{para_big_6_2_multidec}) + 200k steps & 34.2 & 29.0 & 21.2 & .575 & .569 & .484
& .577 & .570 & .484 & 38.1 & 36.7 & 27.5 & 1360 \\ 
		\labelledmodelcounter{para_big_6_2_multidec_multipara} & (\modelref{para_big_6_6_multipara_train_filter}) $\to$ 6-2 & 34.0 & 28.8 & 21.0 & .573 & .566 & .483
& .575 & .568 & .483 & 37.9 & 36.4 & 27.3 & 1372 \\ 
		\labelledmodelcounter{para_big_6_2_multidec_from_6_2} & (\modelref{para_big_6_2}) $\to$ 6-2 & 34.1 & 28.8 & 21.1 & .574 & .566 & .483
& .576 & .568 & .483 & 38.0 & 36.2 & 27.4 & 1295 \\ 
		\labelledmodelcounter{para_big_6_2_multidec_from_6_2_multipara} & (\modelref{para_big_6_2_multipara_train_filter}) $\to$ 6-2 & 34.1 & 28.8 & 21.1 & .574 & .567 & .484
& .576 & .569 & .484 & 37.9 & 36.1 & 27.4 & 1323 \\ 
		\modelref{para_big_12_2_multidec} & (\modelref{para_big_12_2}) $\to$ 12-2 & 35.3 & 29.4 & 21.8 & .583 & .573 & .490
& .585 & .574 & .490 & 39.5 & 37.4 & 28.7 & 1270 \\ 
		\labelledmodelcounter{para_big_12_2_multidec_longer} & (\modelref{para_big_12_2_multidec}) + 200k steps & 35.6 & 29.6 & 22.0 & .585 & .574 & .492
& .587 & .575 & .492 & 39.8 & 37.5 & 28.9 & 1221 \\ 
		\labelledmodelcounter{para_big_12_2_multidec_multipara} & (\modelref{para_big_12_2_multipara_train_filter}) $\to$ 12-2 & 35.2 & 29.3 & 21.8 & .583 & .572 & .491
& .585 & .574 & .491 & 39.5 & 37.2 & 28.6 & 1274 \\ 
        \labelledmodelcounter{para_big_12_2_multidec_freeze} & (\modelref{para_big_12_2_multidec}) + freeze enc. & 35.5 & 29.6 & 20.5 & .584 & .574 & .478
& .587 & .575 & .478 & 39.5 & 37.5 & 26.5 & 1238 \\ 
		\labelledmodelcounter{para_big_12_2_multidec_freeze_multipara} & (\modelref{para_big_12_2_multidec_multipara}) + freeze enc. & 35.1 & 29.5 & 21.6 & .581 & .574 & .489
& .583 & .575 & .489 & 39.2 & 37.3 & 28.3 & 1219 \\ 
		\labelledmodelcounter{para_big_12_2_multidec_no_lang_code} & (\modelref{para_big_12_2_multidec}) w/o lang code & 35.2 & 29.6 & 21.9 & .583 & .573 & .491
& .585 & .575 & .491 & 39.3 & 37.3 & 28.6 & 1215 \\ 
		\labelledmodelcounter{para_big_12_2_multidec_no_lang_code_longer} & (\modelref{para_big_12_2_multidec_no_lang_code}) + 200k steps & 35.5 & 29.8 & 22.1 & .585 & .575 & .493
& .587 & .577 & .493 & 39.7 & 37.7 & \textbf{29.0} & 1242 \\ 
		\labelledmodelcounter{para_big_12_2_multidec_shared_embed} & (\modelref{para_big_12_2_multidec}) + shared embed & 35.2 & 29.5 & 21.8 & .582 & .572 & .489
& .585 & .574 & .490 & 39.3 & 37.2 & 28.4 & 1224 \\ 
	\end{tabular}
	\caption{TED2020 test chrF, FLORES devtest spBLEU \cite{goyal2021flores101} and decoding speed of \textbf{ParaCrawl models} of various depths. WPS: speed in words per second for $\to$EN translation. Scores in parentheses are obtained by pivot translation through English.
	(\modelref{para_wide_12_2}) uses the same architecture as (\modelref{para_big_12_2}) but with a feed-forward dimension of 8192 and is trained for 1M steps on ParaCrawl v8.
	(\modelref{para_big_12_2_multidec_no_lang_code}) is fine-tuned like (\modelref{para_big_12_2_multidec}) but without language codes, similarly to \citet{lyu-etal-2020-revisiting} (while the pre-trained model has language codes). (\modelref{para_big_12_2_multidec_freeze}, \modelref{para_big_12_2_multidec_freeze_multipara}) freeze the shared encoder parameters during the multi-decoder training stage. (\modelref{para_big_12_2_multidec_shared_embed}) is trained with shared target embeddings and $N_{train}=8$k (instead of training one separate embedding matrix per decoder).}
	\label{tab:paracrawl_scores_more}
\end{table}

\end{document}